\documentclass[lettersize,journal]{IEEEtran}
\usepackage[colorlinks,
linkcolor=blue,
anchorcolor=blue,
citecolor=blue
]{hyperref}
\usepackage{amsmath,amsfonts}
\usepackage{algorithm}
\usepackage{algorithmic}
\usepackage{array}
\usepackage[caption=false,font=normalsize,labelfont=sf,textfont=sf]{subfig}
\usepackage{textcomp}
\usepackage{stfloats}
\usepackage{verbatim}
\usepackage{graphicx}
\def\BibTeX{{\rm B\kern-.05em{\sc i\kern-.025em b}\kern-.08em
    T\kern-.1667em\lower.7ex\hbox{E}\kern-.125emX}}
\usepackage{balance}
\usepackage{pifont}
\usepackage{xcolor, color}
\usepackage{booktabs}
\usepackage{makecell}
\usepackage{setspace}
\usepackage{multirow}
\usepackage{ulem}

\usepackage{mathptmx}

\begin{document}
\title{Bi-temporal Gaussian Feature Dependency Guided Change Detection in Remote Sensing Images}


\author{Yi Xiao,~\IEEEmembership{Student Member,~IEEE}, 
	Bin Luo,~\IEEEmembership{Senior Member,~IEEE},
	Jun Liu,
	Xin Su, 
	and Wei Wang 
\thanks{
\IEEEcompsocthanksitem Yi Xiao, Bin Luo, Jun Liu and Wei Wang are with the State Key Laboratory of Information Engineering in Surveying, Mapping and Remote Sensing, Wuhan University, Wuhan 430079, China (e-mail:xy\_whu@whu.edu.cn; luob@whu.edu.cn; liujunand@whu.edu.cn; kinggreat24@whu.edu.cn).
\IEEEcompsocthanksitem Xin Su is with the School of Remote Sensing and Information Engineering,Wuhan University, Wuhan 430079, China (e-mail: xinsu.rs@whu.edu.cn).
\IEEEcompsocthanksitem Corresponding author: Jun Liu.}}


\maketitle

\begin{abstract}
Change Detection (CD) enables the identification of alterations between images of the same area captured at different times. However, existing CD methods still struggle to address pseudo changes resulting from domain information differences in multi-temporal images and instances of detail errors caused by the loss and contamination of detail features during the upsampling process in the network. To address this, we propose a bi-temporal Gaussian distribution feature-dependent network (BGFD). Specifically, we first introduce the Gaussian noise domain disturbance (GNDD) module, which approximates distribution using image statistical features to characterize domain information, samples noise to perturb the network for learning redundant domain information, addressing domain information differences from a more fundamental perspective. Additionally, within the feature dependency facilitation (FDF) module, we integrate a novel mutual information difference loss ($L_{MI}$) and more sophisticated attention mechanisms to enhance the capabilities of the network, ensuring the acquisition of essential domain information. Subsequently, we have designed a novel detail feature compensation (DFC) module, which compensates for detail feature loss and contamination introduced during the upsampling process from the perspectives of enhancing local features and refining global features. The BGFD has effectively reduced pseudo changes and enhanced the detection capability of detail information. It has also achieved state-of-the-art performance on four publicly available datasets - DSIFN-CD, SYSU-CD, LEVIR-CD, and S2Looking, surpassing baseline models by +8.58\texttt{$\%$}, +1.28\texttt{$\%$}, +0.31\texttt{$\%$}, and +3.76\texttt{$\%$} respectively, in terms of the F1-Score metric.
\end{abstract}


\section{Introduction}
\IEEEPARstart{C}{hange} detection aims to detect changes in regions of interest between bi-temporal or multi-temporal images of the same area \cite{rensink2002change}. It is an important research direction in the field of computer vision and has a long history of development in remote sensing, which has a wide application in various fields such as disaster assessment \cite{bovolo2007split}, land use \cite{feranec2007corine}, and ecological environment protection \cite{kennedy2009remote}. 

In the early stages of change detection, manual interpretation was employed to acquire change maps, which was highly inefficient and labor-intensive. Subsequently, traditional change detection utilizing algebraic or transformation methods emerged. For instance, \cite{peli1997multispectral} conducted change detection using a method that compares data statistical distances with a sliding window. Additionally, \cite{celik2009unsupervised} employed principal component analysis (PCA) and k-means clustering method for change detection, enhancing the robustness of the algorithm. However, traditional algorithms may exhibit higher error rates when faced with more complex data and often demonstrate poor generalization capabilities \cite{o2020deep}. 


\begin{figure}[!t]
	\centering
	\includegraphics[width=3.5in]{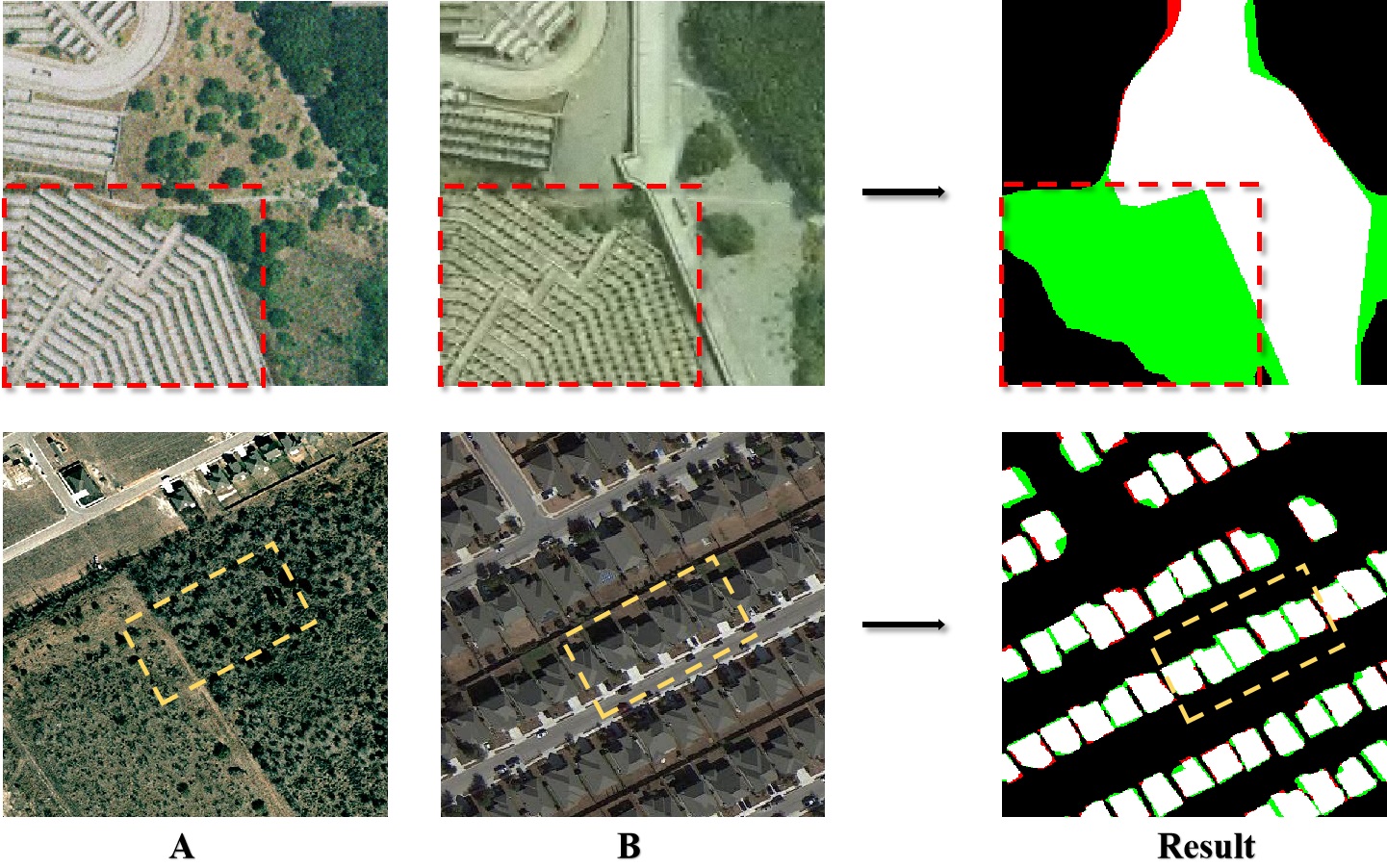}
	\vspace{-1em}
	\caption{The images show the bi-temporal images and the detection results of ChangeCLIP respectively, with green indicating false positive (FP) and red indicating false negative (FN). The top set of images illustrates pseudo changes resulting from illumination difference, while the bottom set of images demonstrates the difficulty in preserving detail of buildings.}
	\label{illumination_error}
\end{figure}

Compared to traditional algorithms, deep learning methods turned out to be very good at discovering intricate structures in high-dimensional data \cite{lecun2015deep}. Researchers have proposed numerous deep learning change detection (DLCD) methods, demonstrating superior detection capabilities even with the use of relatively simple networks such as deep belief networks \cite{argyridis2016building} or autoencoders \cite{chen2016differencing} compared to traditional algorithms. However, in complex and diverse scenarios, the learning and adaptation capabilities of simple networks still fall short of requirements. 

As deep learning continues to advance, the structures of model networks are becoming increasingly complex, leading to enhanced detection abilities. Recent state-of-the-art (SOTA) methods mainly focus on addressing and resolving the following key issues. Firstly, the problem of foreground-background imbalance within change detection datasets. For example, \cite{jiang2023vct} introduced a unique architecture, VcT, different from vision transformer, focusing on learning semantic representations of background information to mitigate the adverse effects of irrelevant changes on the network. Additionally, \cite{han2023hanet} introduced a progressive foreground-balanced sampling strategy allowing the network to more accurately learn features of foreground images in the early stages of training. Secondly, the issue of underutilization of relevant information between bi-temporal images by the network. \cite{fang2023changer} achieved outstanding detection outcomes through a straightforward network architecture and interaction approach, highlighting the significance of bi-temporal image interaction for change detection, while \cite{lin2022transition} emphasized the role of temporal information within bi-temporal images by constructing transitional pseudo videos and leveraging time information to aid in change detection with spatial information. Lastly, enhancing the attention of the network towards change regions, \cite{luo2023multiscale} designed a bidirectional representation of multi-scale feature processing to acquire more detailed change characteristics. \cite{feng2023change} guided the network to focus more on change regions by incorporating attention both before and after feature generation. While the aforementioned methods have shown excellent performance across various datasets, they still struggle to effectively address pseudo changes and detail information loss caused by factors such as illumination and seasonal variations. As shown in Fig.  \ref{illumination_error}, pseudo changes caused by illumination are wrongly identified as change regions, while the preserving of fine edge details also proves to be challenging.


The ChangeCLIP network proposed by \cite{dong2024changeclip} enabled interaction between bi-temporal images through diverse differential features, improving the capability of  the network to extract change features. Additionally, by employing attention mechanisms, the network was directed to focus more on change regions, thus also alleviating the issue of imbalance between background and foreground. Building upon this foundation, ChangeCLIP combined the CLIP model \cite{radford2021learning} to extract semantic features from images and integrated them into the network. The introduction of semantic features indirectly deepened the understanding of domain information and details of objects, thereby partially reducing detection errors caused by pseudo changes and incomplete or missing detail information. However, this approach is only a extrinsically alleviation and still cannot effectively solve the issues. The root causes of these two issues, in our view, are as follows: 1) The essence of pseudo changes are the domain information differences between multi-temporal images, 2) The commonly used upsampling process in networks itself leads to the loss and contamination of detail information.


Addressing the aforementioned two shortcomings, we propose a bi-temporal Gaussian distribution feature-dependent network (BGFD) in this paper. Inspired by \cite{li2021transferable}, we utilize statistical features to characterize domain information, approximate the feature images with a distribution, and sample noise from it to perturb the network, preventing it from learning redundant domain information, thereby reducing pseudo changes caused by domain information differences. To avoid sacrificing essential domain information, we also design a new loss function based on mutual information to constrain the network and employ a more sophisticated attention mechanism \cite{liu2021global} to enhance the adaptability of our network, facilitating the network to generate the feature maps that can correctly using statistical features to represent domain information. To alleviate the issues of detail information loss and contamination during upsampling, we have designed a novel attention specifically for the upsampling process that refines global information and enhances detail information separately for small-scale feature maps, compensating for the contamination and losses in detail information to mitigate error or missed detections of details. 

\begin{figure}[!t]
	\centering
	\includegraphics[width=3.5in]{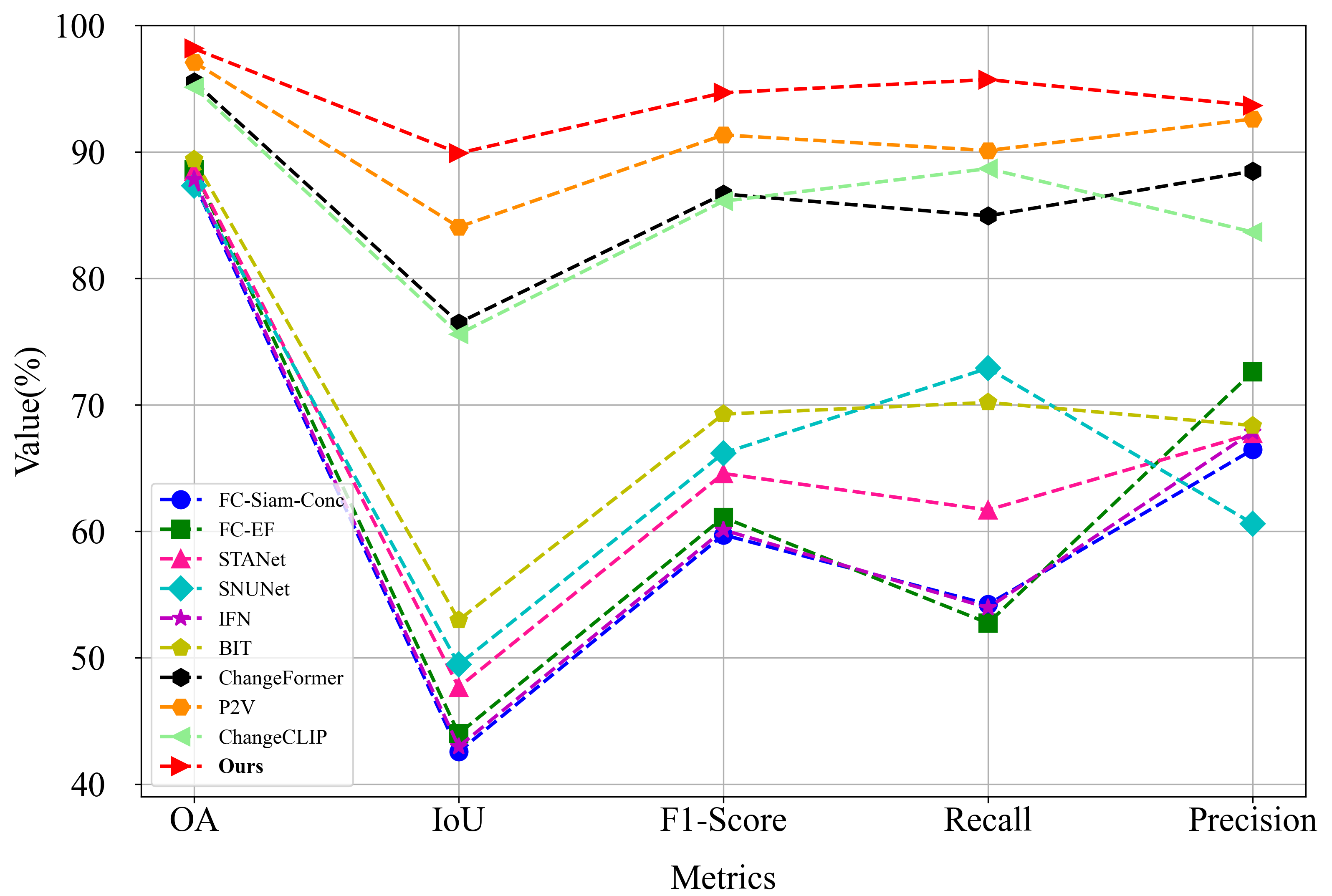}
	\vspace{-1.5em}
	\caption{The line graph illustrating the comparative experimental results on the DSIFN-CD dataset. Our model has been distinctly highlighted in red and enlarged for emphasis. It can be observed that our model achieves the highest position across all five metrics.}
	\label{dsifn_line}
\end{figure}

In summary, the contributions of this paper can be summarized in the following four aspects: 
\begin{enumerate}
\item{We have introduced a novel Gaussian Noise Domain Disturbance (GNDD) module that utilizes distribution to characterize image domain information and samples noise to perturb the network for learning redundant domain information, which fundamentally addresses pseudo changes resulting from domain discrepancies, enhancing the network ability to discern pseudo changes and improving the accuracy of detection results.}
\item{To facilitate the network in producing feature maps with domain statistical features more swiftly, we have developed the Feature Dependency Facilitation (FDF) module integrating mutual information difference loss ($L_{MI}$) and global attention mechanism (GAM) to assist GNDD during the training procedure when the network is not fully trained, further enhancing the change detection capabilities of our network.}
\item{We have also devised a novel attention module named Detail Feature Compensation (DFC), which effectively compensates for the loss and contamination of detail features during the upsampling process from two distinct perspectives of refining global information and enhancing detail information.}
\item{The proposed BGFD has undergone comparative experiments on four public change detection datasets, DSIFN-CD, LEVIR-CD, SYSU-CD, and S2Looking, achieving state-of-the-art results. Here, the line graph of the results on the DSIFN-CD dataset presented in Fig. \ref{dsifn_line} is shown as an example, which demonstrates the effectiveness of our BGFD model.}
\end{enumerate}

The rest of the paper will be organized as follows. Section \uppercase\expandafter{\romannumeral2} provides a description of the relevant work related to DLCD and the utilization of distributions in deep learning. Section \uppercase\expandafter{\romannumeral3}  presents the details of the network we have designed and its core modules. Comparative experiments with other SOTA models and ablation study results are presented in Section \uppercase\expandafter{\romannumeral4}. A conclusion of the entire paper is provided in Section \uppercase\expandafter{\romannumeral5}.

\section{Related Work}
\subsection{RS Image Change Detection Methods}
The aim of remote sensing change detection (RSCD) is to understand the changes between different temporal remote sensing images of the same area \cite{singh1989review}. In the early stages, change detection was achieved through visual analysis and manual interpretation. However, this method was highly inefficient and labor-intensive. For this reason, change detection based on traditional methods began to emerge. For instance, methods like \cite{nelson1983detecting}, \cite{howarth1981procedures} utilized algebraic algorithms, directly using the difference or ratio between images, followed by setting thresholds to identify changed areas. While these algorithms intuitively performed change detection, setting fixed thresholds was challenging. Other traditional methods demonstrated a certain level of adaptability, such as \cite{byrne1980monitoring} implementing PCA for change detection, and \cite{weismiller1977change} adopting a post-classification comparison approach. The emergence of these methods to some extent improved the accuracy of change detection algorithms. However, these traditional algorithms still have drawbacks in terms of generalization, making it difficult to achieve good performance across various datasets. 

With the advancement of computility, change detection algorithms based on deep learning have become viable, offering superior capabilities in deep feature extraction and generalization. In the early stages of DLCD development, various forms of deep learning networks were applied to change detection, such as generative adversarial networks \cite{gong2017generative, jian2021gan}, recurrent neural networks \cite{lyu2016learning, mou2018learning}, stacked autoencoders \cite{gong2017feature}, etc. However, due to the convolutional neural network (CNN) being more suitable for learning hierarchical image representations from input data \cite{yamashita2018convolutional}, CNN was the most widely used in the early DLCD applications \cite{zhang2019detecting, papadomanolaki2021deep, liu2020building, daudt2018fully, ye2023adjacent}. Nevertheless, as CNNs struggle with extracting global features, the emergence of transformer addresseed this issue. Consequently, improved DLCD methods based on the transformer architecture have started to emerge \cite{chen2021remote, bandara2022transformer, lu2022rcdt, zhang2022swinsunet, li2022transunetcd}. These algorithms possess enhanced capabilities in utilizing global features, further enhancing detection accuracy. However, they have not fully leveraged the temporal dependencies between bi-temporal images, making the change detection results susceptible to external influences. Some methods have introduced attention modules to extract spatial-temporal relationships between bi-temporal images, such as \cite{shi2021deeply}, which integrated the convolutional block attention module into deep metric learning methods for change detection, enhancing the learning capabilities of feature extractors. Additionally, \cite{chen2020spatial} designed a new CD self-attention module to capture multi-scale spatial-temporal dependencies, improving the robustness of detection results. However, the methods mentioned above still struggle to effectively address the extraction of detail information and the issue of pseudo changes arising from illumination or seasonal  variations. To tackle these challenges, \cite{chen2023saras} designed a scale-aware attention module to mitigate boundary errors resulting from spatial scaling variations between objects, achieving a certain enhancement in the extraction of detailed boundary information. While \cite{fang2021snunet} designed a densely connected siamese network, which facilitated the precise transmission of shallow spatial information into deep networks, reducing detection errors related to edge pixels and small targets. Moreover, \cite{jia2024siamese, bandara2022ddpm} introduced denoising diffusion probabilistic models \cite{ho2020denoising}, which have recently gained popularity in other tasks, to enhance feature extraction capabilities and improve the robustness of the model to domain information. 

Although these methods partially address these issues, they do not tackle the root causes directly, making it difficult to produce significant effects. The commonly used upsampling process in the model inherently leads to the loss and contamination of details, which is a major hindrance to extracting detail information. Regarding pseudo changes caused by factors like illumination and seasonal variations, these are essentially due to differences in redundant domain information between bi-temporal images. While the distribution characteristics of an image can effectively represent the domain features of the image, better utilization of image distribution characteristics could potentially provide a solution to this problem. In other computer vision tasks, such as object detection, there have been numerous approaches that have incorporated distributional thinking into the field of deep learning, which we will discuss in Section \uppercase\expandafter{\romannumeral2}-B.

\subsection{Distribution In Deep Learning Methods}
To better elucidate the content of this section, we first need to delve into the significance of distributions applied in images in deep learning computer vision. In mathematical terms, distributions generally refer to probability distributions, used to describe the probability patterns of random variable values. For discrete variables, a probability mass function is commonly used for description. The distributions utilized in deep learning methods also represent probability distributions. Specifically in computer vision tasks, they can describe the probability of pixel values appearing at various positions in images. Images portraying specific objects typically obey the specific image distribution patterns of those objects. For example, images containing cats would exhibit pixel value probabilities that follow the distribution pattern characteristic of cat images. 

There are many deep learning methods that directly incorporate distributions to characterize specific features in images. For example, in the task of rotated object detection, \cite{yang2021rethinking} approximated rotated bounding boxes as two-dimensional Gaussian distributions and computed the Wasserstein distance between the predicted boxes and ground truth boxes to address the non-differentiable issue of rotated box losses. In the context of tiny object detection tasks, \cite{xu2022rfla} proposed a Gaussian receptive field-based label assignment strategy, approximating the effective receptive fields of different layers in the network and the ground truth boxes using Gaussian distributions. By minimizing the distance between the distributions, this approach achieved balanced learning for tiny objects. In the realm of data augmentation tasks, \cite{ding2021modeling} employed 3D variational autoencoders to learn the distribution characteristics for labeling medical images. By sampling from the distribution to generate simulated training images, they achieved data augmentation on brain magnetic resonance imaging images. Similarly, \cite{li2021transferable} introduced a transfer semantic augmentation method, in which they measured intra-class semantic differences between the source domain and target domain using multivariate Gaussian distributions. They then utilized the constructed Gaussian distributions for sampling, facilitating infinite augmentation from source to target domain features. 

Inspired by \cite{li2021transferable}, we have integrated the concept of distribution into DLCD. In order to address the issue of pseudo changes in change detection caused by information variations in different domains, we propose BGFD. By fitting domain information in images into distributions within GNDD, we disturb redundant domain information in images by sampling noise from the constructed distribution, enhancing the domain robustness at a more fundamental level.

\begin{figure*}[!t]
	\centering
	\includegraphics[width=7.0in]{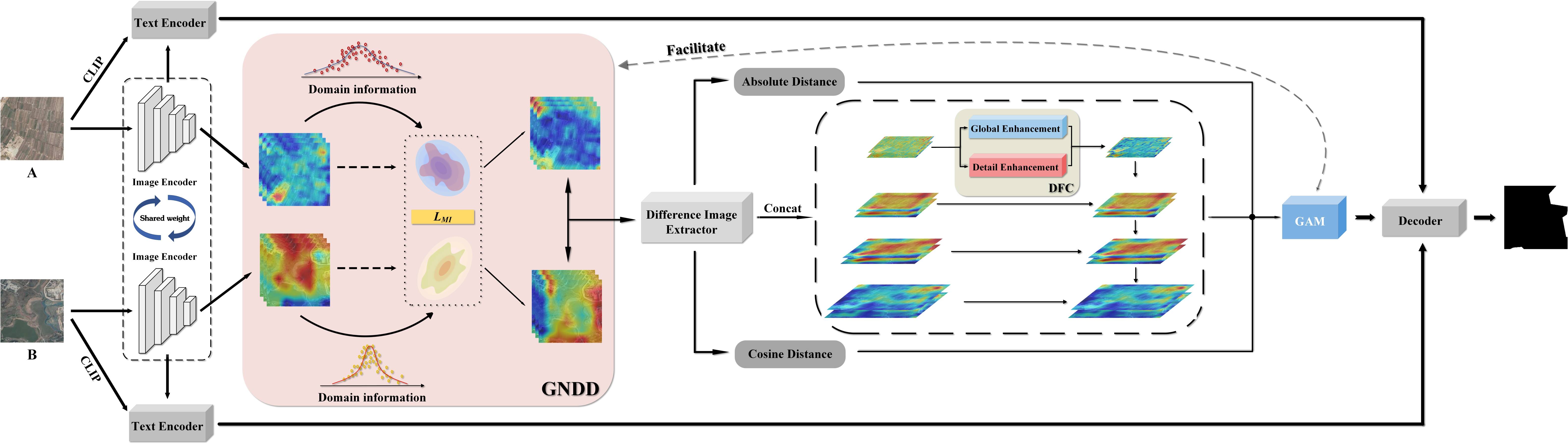}
	\caption{The overall architecture of BGFD. Our innovative modules consist of GNDD, DFC, and FDF module composed jointly by GAM and $L_{MI}$. The network is composed of four main parts. The feature extraction part includes image encoders and text encoders for extracting visual and semantic features of the image, respectively. The Gaussian noise domain disturbance part processes the visual features by approximating distributions and sampling from it to eliminate redundant domain information. The difference image extractor generates three types of difference images, with the concatenated one being processed by the FPN structure, utilizing the DFC module to compensate the detail information. Lastly, A decoder is used to handle semantic features and multiple difference features.}
	\label{XXNet_new}
\end{figure*}

\section{METHODOLOGY}
 In this section, the overall structure of the network is outlined first, followed by DFC module, GNDD module and FDF module.

\subsection{OverView Architecture}

As shown in Fig. \ref{XXNet_new}, BGFD is primarily comprised of four main components: the feature extraction stage, Gaussian noise domain disturbance stage, difference image extractor stage and decoder stage. \textbf{Feature Extraction}: Visual and semantic features were extracted using the ResNet-50 and CLIP models, respectively. \textbf{Gaussian Noise Domain Disturbance}: The visual features was further processed, utilizing the GNDD module to approximate new distributions from the origin distribution of feature maps and sampling noise from it, the redundant domain information was eliminated. Additionally, by calculating the mutual information difference loss between original and new distributions, the training of network was constrained to ensure the learning of essential domain information. \textbf{Difference Image Extractor}: Three forms of difference images were obtained, and when processing the concatenated difference features with the FPN structure, the DFC module was introduced to refine global information and enhance detail information in small-scale feature maps, alleviating detail distortion during upsampling. By concatenating the three forms of difference images and processing them through GAM, the adaptability of our network was enhanced, promoting better functionality of the GNDD module. \textbf{Decoder}: The decoder utilized in the study shared a similar structure with vision-language-driven decoders as used in ChangeCLIP, simultaneously handling semantic features and multiple difference features to obtain detection outcomes.


\subsection{Detail Feature Compensation Module}
\begin{figure}[!t]
	\centering
	\includegraphics[width=3.5in]{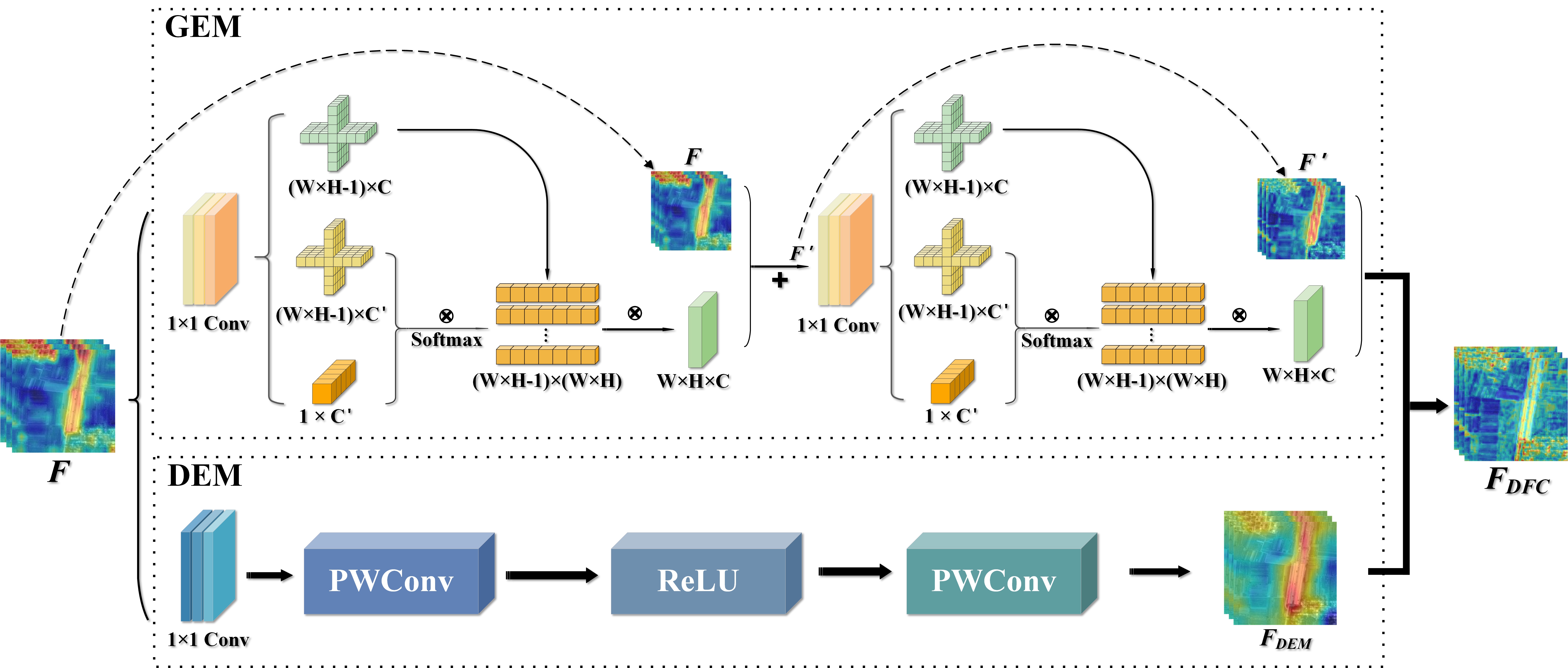}
	\vspace{-1em}
	\caption{Structure of DFC. The DFC module consists of two branches. The GEM branch is used to further refine the global information in small-scale feature maps, while the DEM is responsible for enhancing the detail information in small-scale feature maps.}
	\label{DFC_new}
\end{figure}

The FPN structure is commonly employed for processing and acquiring multi-scale characteristics of images, with the multi-scale features conveying information ranging from global to local levels across various sizes. However, the fusion of small-scale features with large-scale features typically involves the use of linear interpolation, despite small-scale features containing limited local information. This can result in incorrect local information during the interpolation process, leading to two primary issues. Firstly, the introduction of inaccurate local information results in a loss of crucial details inherent in the small-scale feature map itself. Secondly, the interpolated feature maps with erroneous information may contaminate the large-scale feature maps once they are concatenated. While transpose convolution offers a viable solution, considering the problem of computational amount, we introduce a practical module, the detail feature compensation module, designed to address detail loss and contamination. Integrated at the initial stage of the upsampling process within the FPN, the DFC resolves both issues concurrently using two components: the Detail Enhancement Module (DEM) and the Global Enhancement Module (GEM), illustrated in Fig. \ref{DFC_new}.

To counteract the loss incurred by linear interpolation, detail information is amplified by the DEM. Given the limited detail information in small-scale feature maps, we design a straightforward approach to process it to avoid potential distortion caused by excessive processing. Enhancing the details involves obtaining three new feature maps using three $1 \times 1$ convolutions, which are then directly concatenated to obtain $DEM_{head}$. Subsequently, depthwise convolution (dwconv) and pointwise convolution (pwconv) are utilized to restore the original number of channels, resulting in the final $F_{dem}$. The calculation process of DEM is detailed as follows.

\begin{align}
	DEM_{head} =& \mathrm{Concat}(W_{q}(F),W_{k}(F), W_{v}(F))\\
	DEM(F) = \mathrm{pw}&\mathrm{conv}(\mathrm{ReLU}(\mathrm{dwconv}(DEM_{head})))
\end{align}

In order to solve the contamination problem caused by wrong detail information, inspired by the idea of axial attention in \cite{dong2022cswin}, we introduced GEM. In contrast to the DEM, the GEM employs an opposing approach by further diminishing the detail information within the small-scale feature maps. This adjustment aims to minimize the transmission of localized information during the upsampling process, emphasizing global information flow to reduce the introduction of erroneous local details. 


The process involves deriving a new set of $Q$, $K$, and $V$ matrices for the original feature $F$. Subsequently, a deformed cross structure, denoted as $K_{cross}$ $\in \mathbb{R} ^{(W+H-1) \times C \times 1}$, is generated by extending all positions of $K$ both along the channel and vertically and horizontally. By conducting matrix multiplication with the $Q_{cv}$, the channel vector of $Q$ at corresponding positions, $F_{qk} \in  \mathbb{R} ^{(W+H-1)\times W\times H}$ is obtained. Further, a softmax operation is performed on $F_{qk}$ to yield $F_{qk}^{\prime}$. Similarly, all deformed crosses at each position in $V$ are utilized to formulate a new tensor, $V_{cross} \in \mathbb{R} ^ {(W+H-1) \times C \times W \times H}$. Upon matrix multiplication between $F_{qk}^{\prime}$ and $V_{cross}$, the result tensor $F_{gem} \in \mathbb{R} ^{ W \times H \times C}$ is attained. Addition of $F_{gem}$ to $F$ results in $F^{\prime}$, and this process is iterated for $F^{\prime}$ to obtain $F_{GEM}$. The calculation process of GEM unfolds as described above.

\begin{align}
	F_{qk}^{\prime} = Soft&max(K_{cross} \times Q_{cv}) \\
	gem(F) &= V_{cross} \times F_{qk}^{\prime} \\
	GEM(F) &= gem(gem(F))
\end{align}

Finally, we multiply $F_{DEM}$ and $F_{GEM}$ pointwise to obtain $F_{DFC}$.

\begin{align}
	DFC(F) &= DEM(F) \otimes GEM(F)
\end{align}

In this section, we propose a novel attention module, DFC, which addresses the issues of erroneous local information generation and large-scale local information contamination from two distinct perspectives: excluding local information and enhancing the quality of local information, which has led to a higher quality of multi-scale detail information within the traditional FPN structure.

\subsection{Gaussian Noise Domain Disturbance Module}
 The incorporation of DFC has further enhanced the detail information in images. However, pseudo changes induced by domain information such as illumination and seasonal variations cannot be eliminated solely at the level of detail information. Moreover, some pseudo changes manifest precisely at the level of detail. Therefore, solely enhancing pure detail information without considering variations in domain information could potentially exacerbate the impact of pseudo changes. Presently, numerous studies seek to address this issue through attention modules or the inclusion of temporal information. However, while the utilization of attention mechanisms serve as an adaptive means to compensate for domain information, it does not offer a fundamental solution to the problem. while the inclusion of temporal information is actually an attempt to let the network learn and simulate the transition process between domains, which is difficult in the case of insufficient temporal information. In comparison to attention-based solutions and the inclusion of temporal information, the inherent distribution characteristics of features can fundamentally represent the domain information of images. Therefore, by using the distribution pattern of features, we aim to eliminate the impact of pseudo changes caused by inconsistent domain information between bi-temporal images. 

\begin{figure}[!t]
	\centering
	\includegraphics[width=3.5in]{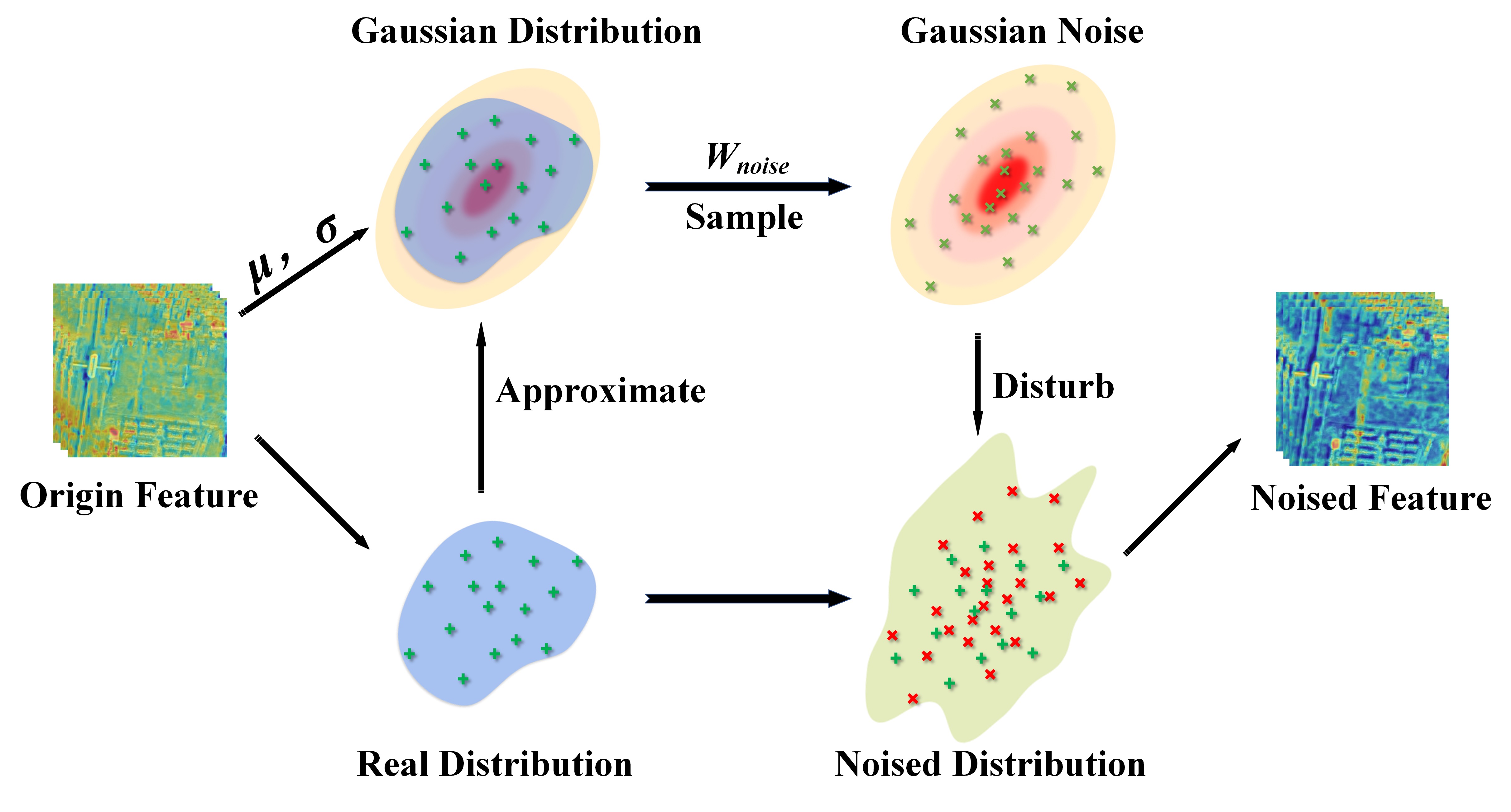}
	\vspace{-2em}
	\caption{Structure of GNDD. The $\mu$, $\sigma$ refer to the mean value and standard deviation of feature map. In the figure, we only illustrate the computation process for feature maps of time A. In practice, we similarly perform calculations on the multi-scale feature maps of time B.}
	\label{GNDD}
\end{figure}

The crux of pseudo changes brought by domain information is the acquisition of redundant domain features. Inspired by \cite{li2021transferable}, our proposed solution is to intentionally perturbing the domain information within the image to constrain the acquisition of domain information. To achieve this, we introduce the Gaussian noise domain disturbance module, shown in Fig. \ref{GNDD}. Initially, we extract the mean and standard deviation from the bi-temporal features outputted by the backbone, establishing a Gaussian distribution based on these parameters to approximate the domain information within the features. The general form of a Gaussian distribution is expressed as follows:
\begin{align*}
	f(x)= \frac{1}{(\sqrt{2\pi} \sigma_{1})}e^{-\frac{1}{2}(\frac{x_{1} - \mu_{1}}{\sigma_{1}})^{2}}\cdots \frac{1}{(\sqrt{2\pi} \sigma_{n})}e^{-\frac{1}{2}(\frac{x_{n} - \mu_{n}}{\sigma_{n}})^{2}}
\end{align*}

Subsequently, we randomly sample noise from the Gaussian distributions and add them to the feature maps to disrupt the domain information learning procedure of the network.

%
\begin{algorithm}
	\caption{Flowchart of GNDD}
	\begin{algorithmic}
		\renewcommand{\algorithmicrequire}{\textbf{Input:}} 
		\REQUIRE \text{Multi-scale bi-temporal features} $F_{t,i}^{s}$ \\ \ \ \ \ \ \ \ \text{output from backbone}, hyperparameter $\lambda$; \\
		\ \ \ \ \ \ \ $f_{t} = 0$, total number of train iters $f_{T}$ \\
		\renewcommand{\algorithmicensure}{\textbf{Output:}}
		\ENSURE \text{noise-disturbed features} $F_{t,i}^{s \prime}$
		
		\ \textbf{for} \text{each $t \in$ [A, B], $s \in$ Feature Scales, $i \in$ Batch} \textbf{do}\\
		\ \ \ \ 1: $ft$ \textbf{+=} $ft;$ \\ 
		\ \ \ \ \ \ \ \text{// Refresh the number of forward propagations} \\
		\ \ \ \ 2: Obtain $\mu_{t,i}^{s}$ = $mean(F_{t,i}^{s})$, \\
		\ \ \ \ \ \ \ $\sigma_{t,i}^{s} = \sqrt{(F_{t,i}^{s} - \mu_{t,i}^{s})^{2}}$ from $F_{t,i}^{s}$; \\
		\ \ \ \ 3: generate Gaussian Distribution \\ 
		\ \ \ \ \ \ \ $g_{t,i}^{s} \sim N(\mu_{t,i}^{s}, \sigma_{t,i}^{s})$; \\
		\ \ \ \ 4: $Noise_{t,i}^{s}=Sample(g_{t,i}^{s}, F_{t,i}^{s}.shape)$; \\
		\ \ \ \ \ \ // Sample Noise tensor from $g_{t,i}^{s}$ with the shape of $F_{t,i}^{s}$ \\
		\ \ \ \ 5: C = min($f_{t} / f_{T}$, 1.0); \\ 
		\ \ \ \ 6: $w_{noise} = C \times \lambda $; // Get the noise weight\\
		\ \ \ \ 7: $F_{t,i}^{s\prime} = w_{noise}\times Noise_{t,i}^{s} + F_{t,i}^{s}$; \\
		\ \ \ \ // Add noise to original feature maps \\
		\ \textbf{end for}
	\end{algorithmic}
	\label{algorithm}
\end{algorithm}

The GNDD algorithm procedure is shown in Algorithm \ref{algorithm}. In this study, the feature sizes obtained from backbone are {64, 32, 16, 8}. To facilitate a better distinction between the Gaussian distribution, sampled noise, and feature images of the bi-temporal multi-scale images, we employ subscripts denoted by lowercase $t$ to indicate the image time label, $i$ to represent the image number in the batch, and superscripts indicated by lowercase $s$ to denote the image scale. Since the image domain information in the same batch may not be consistent, mean ($\mu_{t,i}^{s}$) and standard deviation ($\sigma_{t,i}^{s}$) values are calculated separately for each image in the batch as follows.
\begin{equation}
	\mu_{t,i}^{s} = \sum\limits_{p}^{P}\frac{F_{t,i}^{s}(p)}{N} 
\end{equation}
\begin{equation}
	\sigma_{t,i}^{s} = \sqrt{\sum\limits_{p}^{P}\frac{(F_{t,i}^{s}(p)-\mu_{t,i}^{s})^{2}}{N}}
\end{equation}


\noindent where \textit{N} is number of pixels, \textit{F} are the feature maps sent into GNDD, and \textit{L} is the position on the feature maps, respectively.

Consequently, a series of Gaussian distributions $G={g_{t,i}^{s}}$ are generated with a total number of $batch \times t_{num} \times scale_{num}$, where $t_{num}$ representing the total number of time phases (2 in this study) and $scale_{num}$ representing the total number of feature scales (4 in this study). Subsequently, noise $Noise_{t,i}^{s}$ is randomly sampled from each Gaussian distribution $g_{t,i}^{s}$ and the sampling process can be simply represent as follows.

\begin{equation}
	Noise_{t,i}^{s}=\mu_{t,i}^{s} + \sigma_{t,i}^{s} \cdot Z
\end{equation}
\noindent where $Z$ is the collection of random numbers sampled from a standard normal distribution.

Then the $Noise_{t,i}^{s}$ is added to the corresponding feature $F_{t,i}^{s}$ acquired from the backbone. The resulting $F_{t,i}^{s\prime}$ is then passed on to the subsequent module for further processing.
\begin{equation}
	F_{t,i}^{s\prime}=F_{t,i}^{s} + Noise_{t,i}^{s}
\end{equation}

To prevent overwhelming the network with excessive noise during the initial training phase, a training noise scheduler was devised. A dynamic adjustment factor $c$ that varies during the training process was first designed, combined with a new hyperparameter $\lambda$ , resulting in a weight $w_{noise}$ that controls the noise. The training noise scheduler is outlined as follows.

\begin{equation}
	C = \min(f_{t} / f_{T}, 1.0)
\end{equation}
\begin{equation}
	w_{noise} = C \times \lambda
\end{equation}

\noindent where $f_{t}$ denotes the number of forward propagations during the training process, $f_{T}$ represents the total iteration number of training, $\lambda$ is 1.0 in this paper. 

In this section, we introduce the GNDD, which aims to enhance the robustness of the network in detecting variations among images captured at different domains. We approximate feature maps as Gaussian distributions and intentionally introduce noise disruption to distract network from learning redundant domain features. By leveraging a distribution-based approach, we fundamentally alleviate the impact of pseudo changes caused by inconsistent domain information in bi-temporal images, thereby improving the detection robustness across images from varying domains.


\begin{figure}[!t]
	\centering
	\includegraphics[width=3.5in]{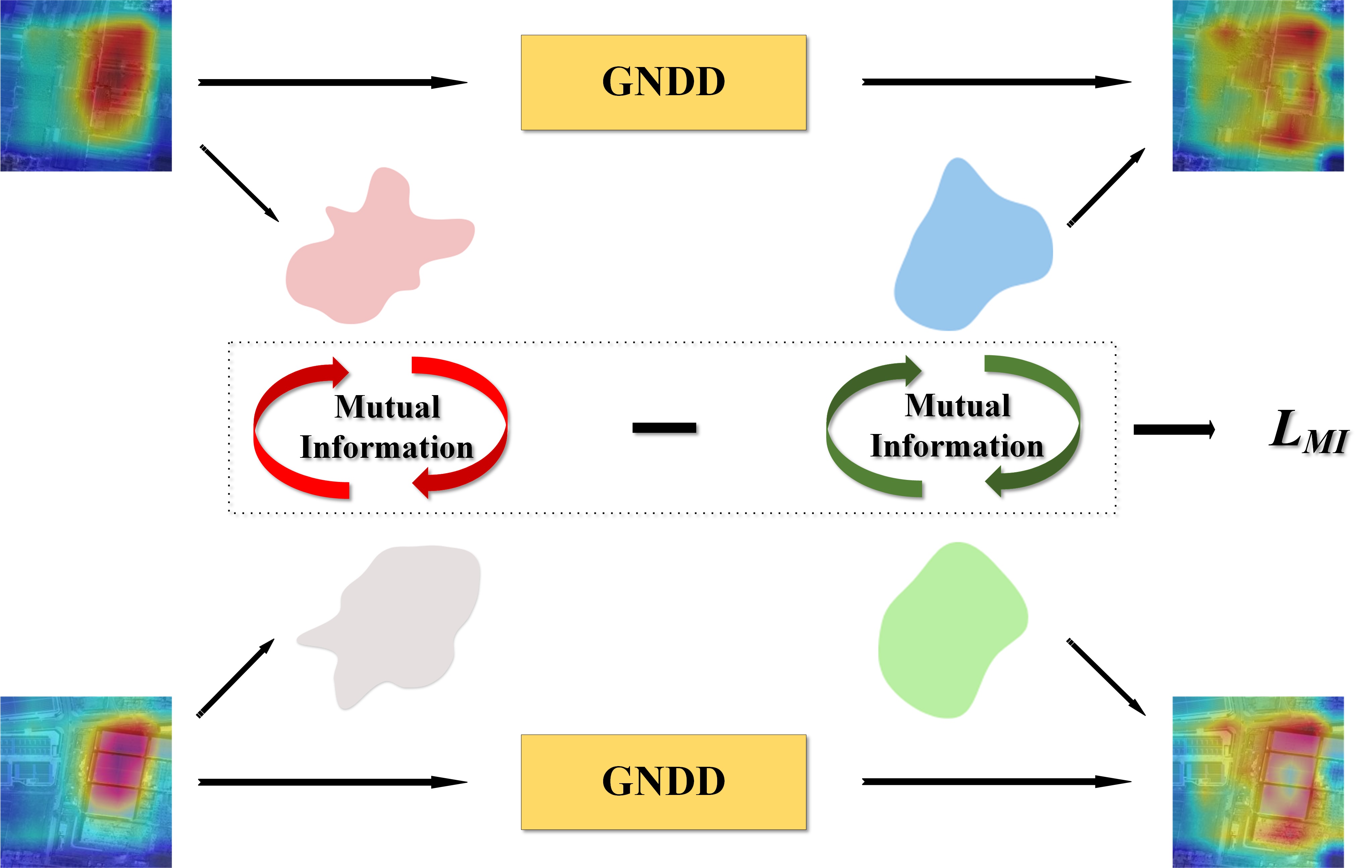}
	\vspace{-1em}
	\caption{A brief schematic diagram of $L_{MI}$, which illustrates the process of calculating the mutual information between feature maps before and after GNDD module, followed by the computation of their differences.}
	\label{MILoss}
\end{figure}

\subsection{Feature Dependency Facilitation Module}
As mentioned above, the purpose of adding Gaussian distributed noise to the feature maps is to disrupt the learning of redundant domain information through interference in domain information learning. However, during the training process, since the network has not completed learning, the statistical information of the generated feature maps may not effectively represent the domain information of the original images, which may cause the approximation of the feature map results in an incorrect Gaussian distribution, leading to erroneous domain information and Gaussian noise interference, adversely affecting the network training. Therefore, to enable the GNDD to function as intended, we have incorporated the feature dependency facilitation module, which consists of two parts. We first introduced mutual information to measure the similarity of information between bi-temporal images. Then, we compared the mutual information of the feature maps before and after adding noise to form $L_{MI}$, ensuring that Gaussian noise correctly interferes with redundant domain information. Additionally, by employing a relatively more complex attention mechanism, we can more efficiently utilize multi-temporal difference images, making it easier for the network to converge to the desired feature maps. 

After adding corresponding Gaussian noise to the bi-temporal images, theoretically, the domain information distance between them will decrease. Since the bi-temporal images depict the same region, removing redundant domain information will increase the interdependence of information between the two feature maps. Initially, we calculate the information entropy \textit{H(T1)}, \textit{H(T2)}, and joint entropy \textit{H(T1, T2)} of the bi-temporal feature maps, measuring the information contained in the feature maps individually and the total shared information between the two feature maps. The mutual information represents the overall interdependence of information between \textit{T1} and \textit{T2}, which we denote as \textit{I(T1; T2)}. To ensure an increase in mutual information between the bi-temporal features after the noise is added, we introduce $L_{MI}$. A brief diagram of $L_{MI}$ is shown as Fig. \ref{MILoss}. Assuming the mutual information of the bi-temporal images before adding noise is $I_{former}$, and after adding noise is $I_{latter}$, the calculation method of $L_{MI}$ is expressed by the following formula. 

\begin{align*}
	H(T)&=\sum\limits_{l \in L}p_{T}(l)\log\frac{1}{p_{T}(l)} \\
	H(T_{1}, T_{2})&=\sum\limits_{l_{1} \in L_{1}}\sum\limits_{l_{2} \in L_{2}}p(l_{1}, l_{2})\log\frac{1}{p(l_{1}, l_{2})} \\
	I(T_{1}; T_{2})&=H(T_{1}) + H(T_{2}) - H(T_{1}, T_{2})
\end{align*}

\begin{equation*}
	L_{MI} = \begin{cases}
		{I_{former} - I_{latter}},&{\text{if}}\ {(I_{latter} - I_{former}) < 0} \\ 
		{\ \ \ \ \ \ \ \ \ \ 0 \ \ \ \ \ \ \ \ \ ,}&{\text{otherwise.}} 
	\end{cases}
\end{equation*}

\noindent where $p$ represents the joint probability distribution function of two discrete random variables $X$ and $Y$, while $p(x)$ and $p(y)$ are the marginal probability distribution functions of $X$ and $Y$, respectively. 



We further enhance the adaptive attention to changing regions to facilitate the acquisition of desired feature maps. Our baseline model, ChangeCLIP, utilizes various forms of differential images to boost the capability of network. However, simple attention mechanisms such as squeeze-and-excitation may not fully leverage the diversity of differential images. To address this, we employ a relatively more complex global attention mechanism to handle the different forms of difference images. This allows the network to adaptively focus on changing regions, thereby enhancing its capacity and promoting the learning of feature maps that can represent domain information through statistical features. Additionally, this facilitates the intended functionality of our GNDD. 

The structure of GAM is achieved through the concatenation of channel attention and spatial attention. The channel attention utilizes a simple multi-Layer perceptron structure, while the spatial attention is carried out through two convolutional layers. The result obtained from each attention operation is element-wise multiplied with the original feature map. The computational process is as follows in the equation provided. 

\begin{align}
	M_{c}(F) = \mathrm{FC_{2}}(&\mathrm{ReLU}(\mathrm{FC_{1}}(F))) \otimes F \\
	M_{s}(F) = \mathrm{BN}(\mathrm{Conv_{2}}(\mathrm{Re}&\mathrm{LU}(\mathrm{BN}(\mathrm{Conv_{1}}(F))))) \otimes F \\
	GAM(F) =& \mathrm{Conv_{3}}(M_{s}(M_{c}(F)))
\end{align}

\noindent where $M_{c}$ is channel attention module in GAM, $M_{s}$ is spatial attention module in GAM, BN is batch normalization layer and FC is fully connection layer. 

By promoting an increase in feature dependency before and after noise injection and enhancing network capability through more sophisticated attention mechanism, the FDF module facilitates the training of feature maps that can represent domain information through statistical features as desired, ensuring the intended functionality of the GNDD.

\section{EXPERIMENTS}
\subsection{Datasets}
In this article, we have employed a total of four open-source datasets, including DSIFN-CD, SYSU-CD, LEVIR-CD and S2Looking. Next, we provide a brief introduction to four datasets, and the key information of the datasets is presented in Table \ref{tab1}.

\begin{table}
	\renewcommand{\arraystretch}{1.25}
		\centering
		\caption{DATASET INTRODUCTION.}
		\label{tab1}
		\begin{tabular}{ c | c  c  c  c }
			\Xhline{1.2pt}
			\textbf{Dataset} & \textbf{Image Size} & \textbf{Train} & \textbf{Validation} & \textbf{Test}\\
			\hline
			DSIFN-CD & 256 $\times$ 256 & 14400 & 1360 & 199\\
			SYSU-CD & 256 $\times$ 256 & 12000 & 4000 & 4000\\ 
			LEVIR-CD & 1024 $\times$ 1024 & 445 & 64 & 128\\
			S2Looking & 1024 $\times$ 1024 & 3500 & 500 & 1000\\
			\Xhline{1.2pt}
		\end{tabular}
		\vspace{-1em}
\end{table}

\subsubsection{DSIFN-CD}
\textit{DSIFN-CD} \cite{zhang2020deeply} is a collection of change detection data obtained from Google Earth, covering six cities in China. The original dataset was uniformly cropped to size of 256$\times$256 based on the distribution of the cropping data in the original dataset. 

\subsubsection{SYSU-CD}
\textit{SYSU-CD} is a high-resolution aerial images (0.5m, pixel) change detection dataset collected in Hong Kong from 2007 to 2014 with a variety of change types. We followed the official distribution method of the dataset. 

\subsubsection{LEVIR-CD}
\textit{LEVIR-CD} is a large remote sensing dataset for building change detection, composed of very high-resolution (0.5m, pixel) Google Earth image patches,capturing a time span of 5-14 years. We followed the official distribution method of the dataset. 

\subsubsection{S2Looking}
\textit{S2Looking} \cite{shen2021s2looking} spans from 2017 to 2020, comprising large-scale side-looking satellite images (0.5-0.8m, pixel) captured by diverse satellites at various off-nadir angles. We followed the official distribution method of the dataset. 


\subsection{Implementation Details}
The proposed BGFD was trained in the PyTorch environment utilizing GeForce RTX 3090 GPU, with a total of 60,000 iterations. The AdamW optimizer was employed, commencing with an initial learning rate of 0.00014 and a weight decay of 0.01.

During training, images exceeding the size of 256$\times$256 were initially divided into patches and then merged, the stride to crop the images is 128 $\times$ 128, with a batch size set at 22. Before feeding the images into the network, the images are normalized with a mean of [123.675, 116.28, 103.53] and a standard deviation of [58.395, 57.12, 57.375]. Taking computational complexity into consideration, we only calculate the $L_{MI}$ using the features from the smallest scale, specifically the first 128 channels of the 8$\times$8 scale feature map.

\begin{figure}[!t]
	\centering
	\includegraphics[width=3.5in]{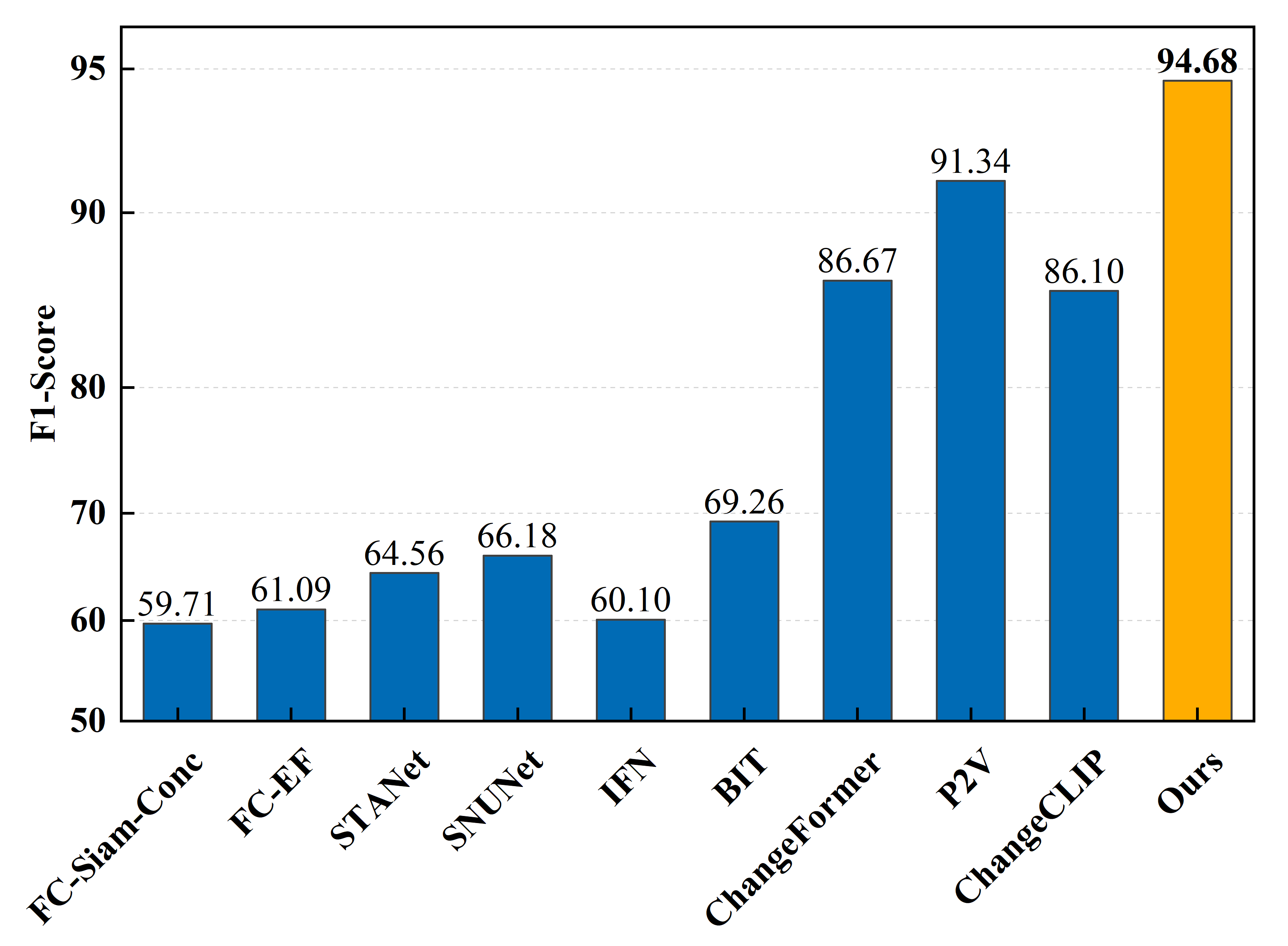}
	\vspace{-1em}
	\caption{Comparison experiment results on DSIFN-CD.}
	\vspace{-1em}
	\label{DSIFN}
\end{figure}

\subsection{Evaluation Metrics}
In order to comprehensively and accurately evaluate the performance of our model on the dataset, a total of 5 metrics were employed, namely Overall Accuracy (OA), Intersection over Union (IoU), F1-Score (F1), Recall (Rec.), and Precision (Prec.). The formulas defining these metrics are as follows:

\begin{align}
	\mathrm{Precision} &= \mathrm{\frac{TP}{TP+FP}} \\
	\mathrm{Recall} &= \mathrm{\frac{TP}{TP+FN}} \\
	\mathrm{IoU} &= \mathrm{\frac{TP}{TP+FN+FP}} \\
	\mathrm{OA} &= \frac{\mathrm{TP}+\mathrm{TN}}{\mathrm{TP+TN+FN+FP}} \\
	\mathrm{F1} &= \mathrm{(1+\beta^{2}) \frac{Precision \times Recall}{\beta ^{2} (Precision + Recall)}}
\end{align}

\begin{table*}
		\centering
		\caption{Quantitative Results On The \textbf{Dsifn-cd} Dataset and \textbf{Sysu-cd} Dataset. The Best And Second Results Are Marked In \textbf{Bold} And \uline{Underlined}, Respectively. All These Scores Are Written In Percentage (\%)}
		\begin{tabular}{ccccccccccccc}
			\specialrule{1.2pt}{1pt}{1.5pt}
			\multirow{2}{*}{\textbf{Methods}}  & \multicolumn{5}{c}{\textbf{DSIFN-CD}} & \multirow{7}{*}{} & \multirow{2}{*}{\textbf{Methods}} & \multicolumn{5}{c}{\textbf{SYSU-CD}} \\ 
			\cmidrule(r){2-6} \cmidrule(r){9-13}  
			\textbf{} & OA & IoU & F1 & Rec. & Prec. & &\textbf{} & OA & IoU & F1 & Rec. & Prec. \\
			\specialrule{1.2pt}{1pt}{1pt}
			FC-Siam-Conc  & 87.57  & 42.56  & 59.71  & 54.21  & 66.45 &
			&  FC-Siam-Conc  & \textbackslash  & 61.75  & 76.35  & 71.03  & 82.54  \\
			FC-EF 		  & 88.59  & 43.98  & 61.09  & 52.73  & 72.61 &
			& DSAMNet 	  & \textbackslash  & 64.18  & 78.18  & 81.86  & 74.81   \\
			STANet  	  & 88.49  & 47.66  & 64.56  & 61.68  & 67.71 & 
			& STANet  	  & \textbackslash  & 63.09  & 77.37  & \textbf{85.33}  & 70.76   \\
			SNUNet  	  & 87.34  & 49.45  & 66.18  & 72.89  & 60.60 &
			& SNUNet  	  & 90.79  		 & 66.02  & 79.54  & 75.87  & 83.58 \\
			IFN  		  & 87.83  & 42.96  & 60.10  & 53.94  & 67.86 & 
			& L-UNet  	  & 90.58  		 & 66.15  & 79.63  & 78.08  & 81.24  \\
			BIT  		  & 89.41  & 52.97  & 69.26  & 70.18  & 68.36 & 
			& BIT 		  & \textbackslash  & 64.59  & 81.62  & 78.49  & 85.00  \\
			ChangeFormer  & 95.56  & 76.48  & 86.67  & 84.94  & 88.48 & 
			& ChangeFormer  & 91.85  		 & 69.46  & 81.98  & 78.57  & 85.69 \\
			P2V  		  & \uline{97.10}  & \uline{84.06}  & \uline{91.34}  & \uline{90.11}  & \uline{92.60} & 
			& VcT  		   & 91.38  		 & 68.22  & 81.11  & 78.44  & 83.97  \\
			ChangeCLIP    & 95.13  & 75.59  & 86.10  & 88.68  & 83.66 & 
			& ChangeCLIP    & \uline{92.22}  		 & \uline{70.91}  & \uline{82.98}  & 80.40  & \uline{85.74} \\
			\specialrule{1.2pt}{1pt}{1.5pt}
			\textbf{BGFD}        & \textbf{98.17}       & \textbf{89.90}        & \textbf{94.68}       & \textbf{95.72}        & \textbf{93.67} & 
			& \textbf{BGFD}        & \textbf{92.78}   & \textbf{72.80}   & \textbf{84.26}       & \uline{82.00}        & \textbf{86.65} \\
			\Xhline{1.2pt}
		\end{tabular}
		\label{tab2}
	\end{table*}


\begin{figure*}[!t]
	\centering
	\includegraphics[width=7in]{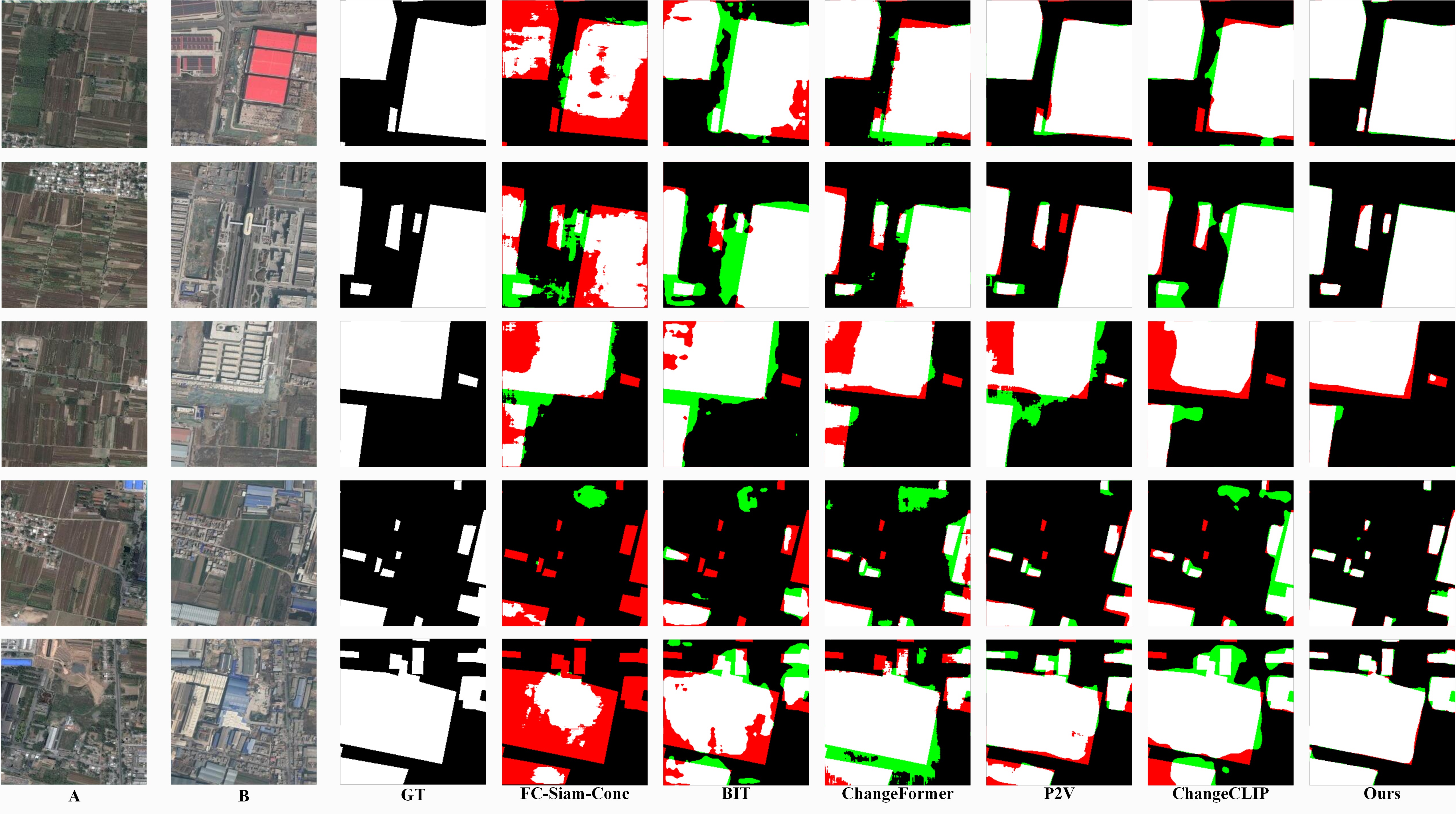}
	\caption{Local visualization of different methods on the DSIFN-CD. True positives (TP) are represented by white pixels, while true negatives (TN) are black. False positives (FP) are green and false negatives (FN) are red.}
	\label{dsifn_result}
\end{figure*}

\begin{figure}[!t]
	\centering
	\includegraphics[width=3.5in]{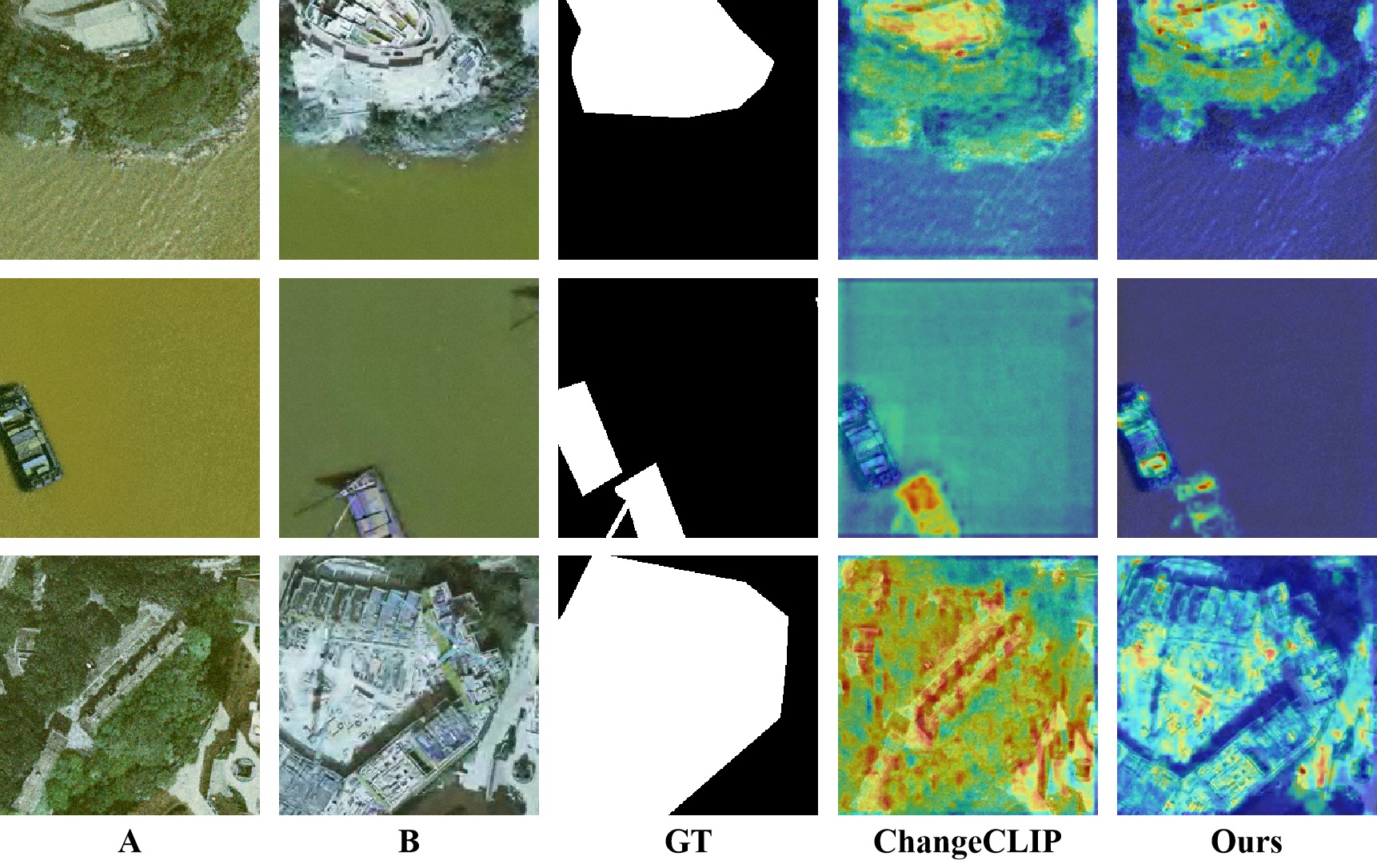}
	\vspace{-1em}
	\caption{Heatmap of visualization results on the DSIFN-CD.}
	\label{dsifn_feature_compare}
\end{figure}

\subsection{Comparison Experiments}

To validate the effectiveness of BGFD in remote sensing image change detection task, we compared it with existing SOTA methods on four datasets. The selected methods include FC-Siam-Conc, FC-Siam-Diff, FC-EF, CDNet, DTCDSCN, DSAMNet, IFN, L-UNet, STANet, SNUNet-CD, VcT, BIT-CD, ChangeFormer, P2V-CD, and ChangeCLIP.

\subsubsection{DSIFN-CD}
In this dataset, we compared our model with 10 methods, and the experimental results are shown in Table \ref{tab2}. The best results are highlighted in bold, and the second-best results are marked in underlined. Additionally, we created a more intuitive histgram to showcase the performance of each method on the dataset, as depicted in Fig. \ref{DSIFN}. On this dataset, we achieved 98.17\% Overall Accuracy, 89.90\% IoU, 94.68\% F1-Score, 95.72\% Recall, and 93.67\% Precision, outperforming our baseline model, ChangeCLIP, by +3.04\% in Overall Accuracy, +14.31\% in IoU, +8.58\% in F1-Score,  +7.04\% in Recall and +10.01\% in Precision, respectively. We presented comparative detection result images of various methods on the DSIFN-CD dataset, as shown in Fig. \ref{dsifn_result}. It is apparent that in our results, the number of false positives has significantly decreased, and the capability in capturing the edge of objects and minor changes have notably improved. This exceptional performance is attributed to three innovative modules proposed by us. The GNDD module enhances the robustness of our network to domain information differences by blocking the learning of redundant domain information, markedly suppressing the generation of pseudo changes caused by variations in illumination, seasons and other factors. Additionally, the FDF module constrains the training process of our network through $L_{MI}$, further boosting the capabilities of the GNDD module. It also utilizes more complex attention mechanism to enable the network to focus more on change areas, enhancing its adaptability to complex scenarios. Lastly, our DFC module compensates the loss and contamination of detail information during upsampling, making it easier for the network to capture building edges and small change regions. These three modules collectively led to improvements in Recall and Precision, consequently raising the evaluation metrics of Overall Accuracy, F1-Score, and IoU.


In Fig. \ref{dsifn_feature_compare}, we compared the feature maps of our proposed BGFD with those of the baseline model ChangeCLIP. It is evident from the heatmaps that our network exhibits enhanced robustness against pseudo-variations. Notably, the network remains unaffected by illumination variations in water surface areas, focusing its attention more on the regions that exhibit changes. Furthermore, the feature heatmaps indicate that our network is capable of extracting clearer and more comprehensive detail information. Although the overall color intensity of our heatmap is darker than that of the baseline model, the contrast in color between the regions of change and unchange is significantly higher. 


\subsubsection{SYSU-CD}
\begin{figure}[!t]
	\centering
	\includegraphics[width=3.5in]{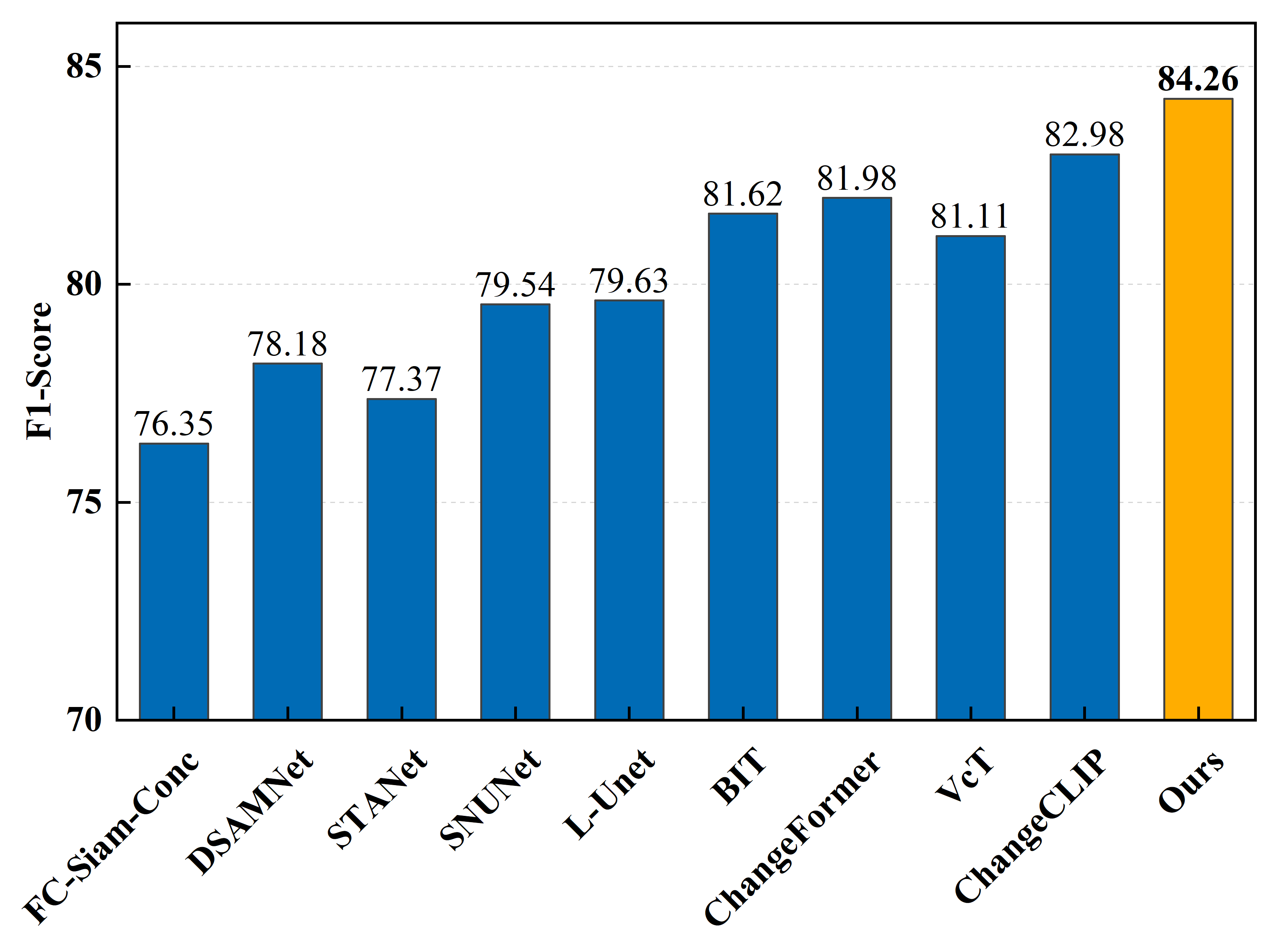}
	\vspace{-1em}
	\caption{Comparison experiment results on SYSU-CD.}
	\label{SYSU-CD}
\end{figure}

We compared our model with 10 selected comparative methods on this dataset, and the experimental results and histgram can be found in Table \ref{tab2} and Fig. \ref{SYSU-CD}, respectively. On this dataset, our model achieved 92.78\% Overall Accuracy, 72.80\% IoU, 84.26\% F1-Score, 82.00\% Recall, and 86.65\% Precision, while our baseline model obtained 92.22\% Overall Accuracy, 70.91\% IoU, 82.98\% F1-Score, 80.40\% Recall, and 85.74\% Precision, resulting in improvements of +0.56\%, +1.89\%, +1.28\%, +1.60\%, and +0.91\%, respectively. 

\begin{figure}[!t]
	\centering
	\includegraphics[width=3.5in]{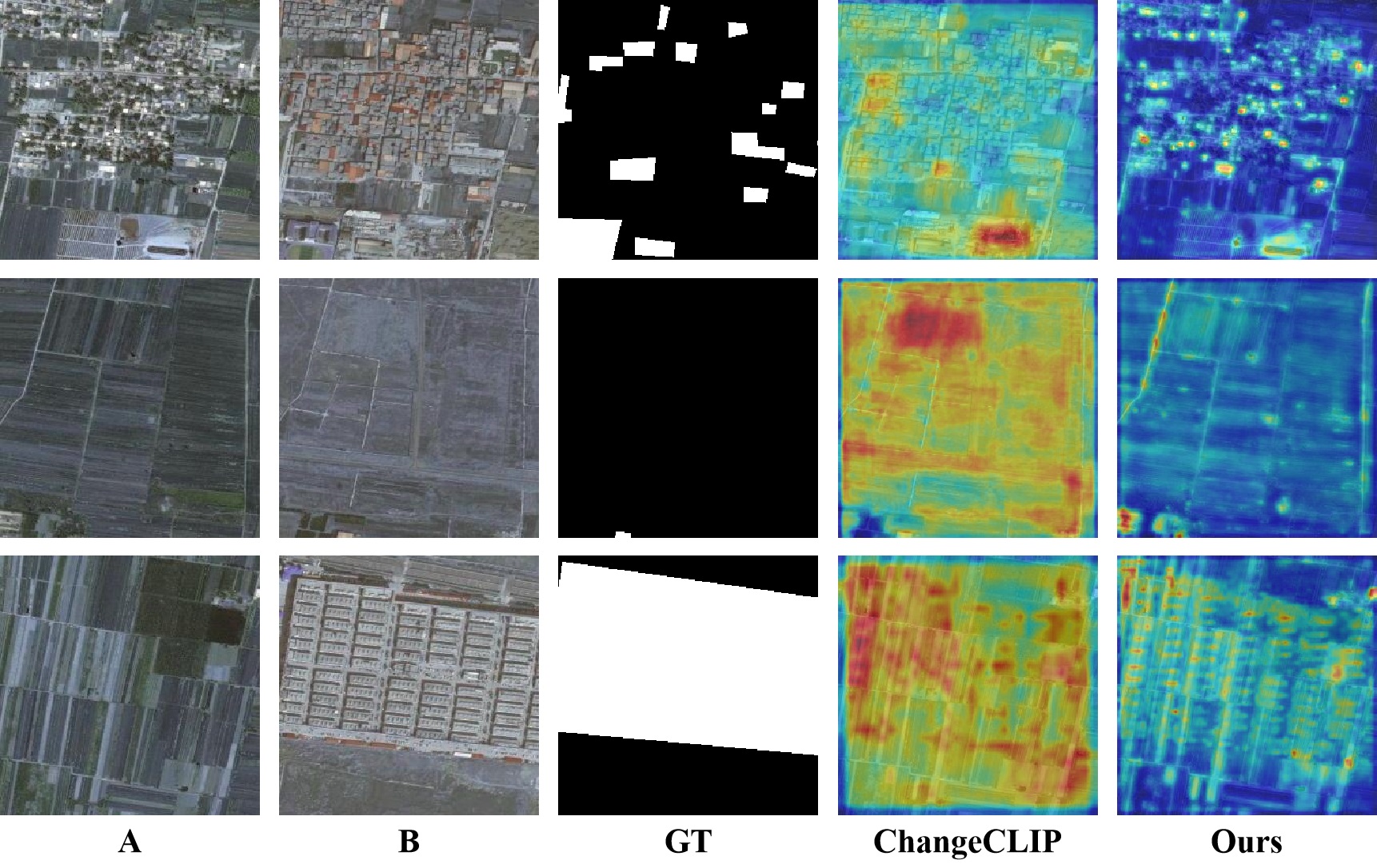}
	\vspace{-1em}
	\caption{Heatmap of visualization results on the SYSU-CD.}
	\label{sysu_feature_compare}
\end{figure}

\begin{figure*}[!t]
	\centering
	\includegraphics[width=7in]{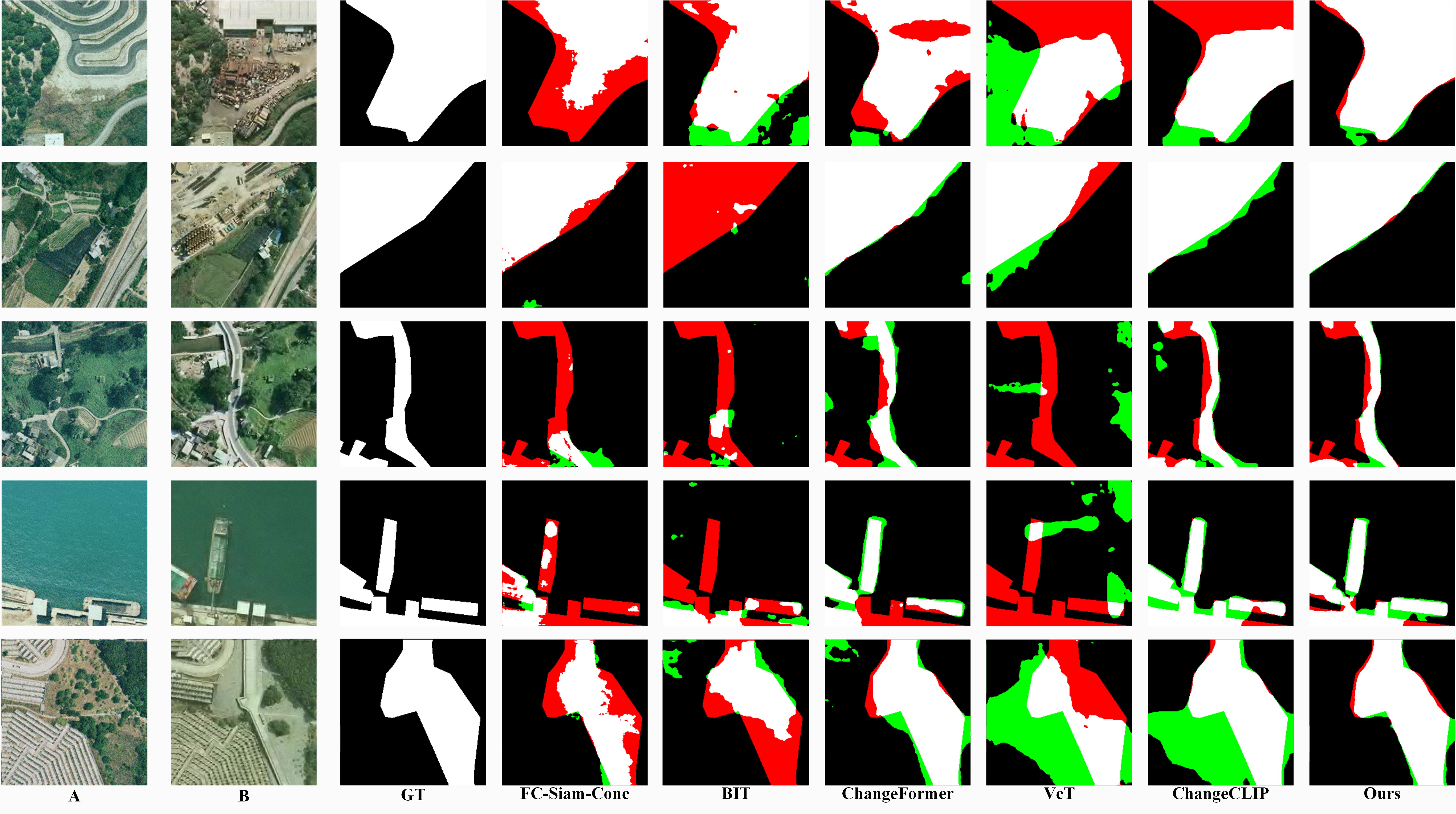}
	\caption{Local visualization of different methods on the SYSU-CD. True positives (TP) are represented by white pixels, while true negatives (TN) are black. False positives (FP) are green and false negatives (FN) are red.}
	\label{sysu_result}
\end{figure*}

Furthermore, we compared the heatmaps of feature maps generated by our model with those of ChangeCLIP on this dataset, as shown in Fig. \ref{sysu_feature_compare}, where our model also demonstrate great ablities of capturing the details and identifying pesudo changes, which is similar to the performance on DSIFN-CD. Additionally, Fig. \ref{sysu_result} displays the detection results of various methods on the SYSU-CD dataset, further confirming the capabilities of our model in terms of detail representation and pseudo-change exclusion.

\begin{table}
	\centering
	\caption{Quantitative Results On The \textbf{Levir-cd} Dataset. The Best And Second Results Are Marked In \textbf{Bold} And \uline{Underlined}, Respectively. All These Scores Are Written In Percentage (\%)}
	\begin{tabular}{cccccc}
		\specialrule{1.2pt}{1pt}{1.5pt}
		\multirow{2}{*}{\textbf{Methods}}  & \multicolumn{5}{c}{\textbf{LEVIR-CD}} \\
		\cmidrule(r){2-6}
		\textbf{}  & OA & IoU & F1 & Rec. & Prec. \\
		\specialrule{1.2pt}{1pt}{1pt}
		FC-Siam-Conc  & 98.49  & 71.96  & 83.69  & 76.77  & 91.99  \\
		DTCDSCN	   & 98.77  & 78.05  & 87.67  & 86.83  & 88.53  \\
		STANet  	   & 98.66  & 77.40  & 87.26  & \textbf{91.00}  & 83.81  \\
		SNUNet 	   & 98.82  & 78.83  & 88.16  & 87.71  & 89.18  \\
		L-UNet  	   & 90.58  & 66.15  & 79.63  & 78.08  & 81.24  \\
		BIT  		   & \textbf{99.57}  & 83.94  & 91.27  & 90.66  & 91.88  \\
		ChangeFormer  & 99.04  & 82.48  & 90.40  & 88.80  & 92.05  \\
		P2V  		   & \textbackslash  & \textbackslash  & \uline{91.94}  & 90.60  & \uline{93.32}  \\
		ChangeCLIP    & 99.17  & \uline{84.77}  & 91.76  & 90.28  & 93.28  \\
		\specialrule{1.2pt}{1pt}{1.5pt}
		\textbf{BGFD}        & \uline{99.20}       & \textbf{85.30}        & \textbf{92.07}       & \uline{90.77}        & \textbf{93.41}   \\
		\Xhline{1.2pt}
	\end{tabular}
	\label{tab3}	
\end{table}

\subsubsection{LEVIR-CD}
\begin{figure}[!t]
	\centering
	\includegraphics[width=3.5in]{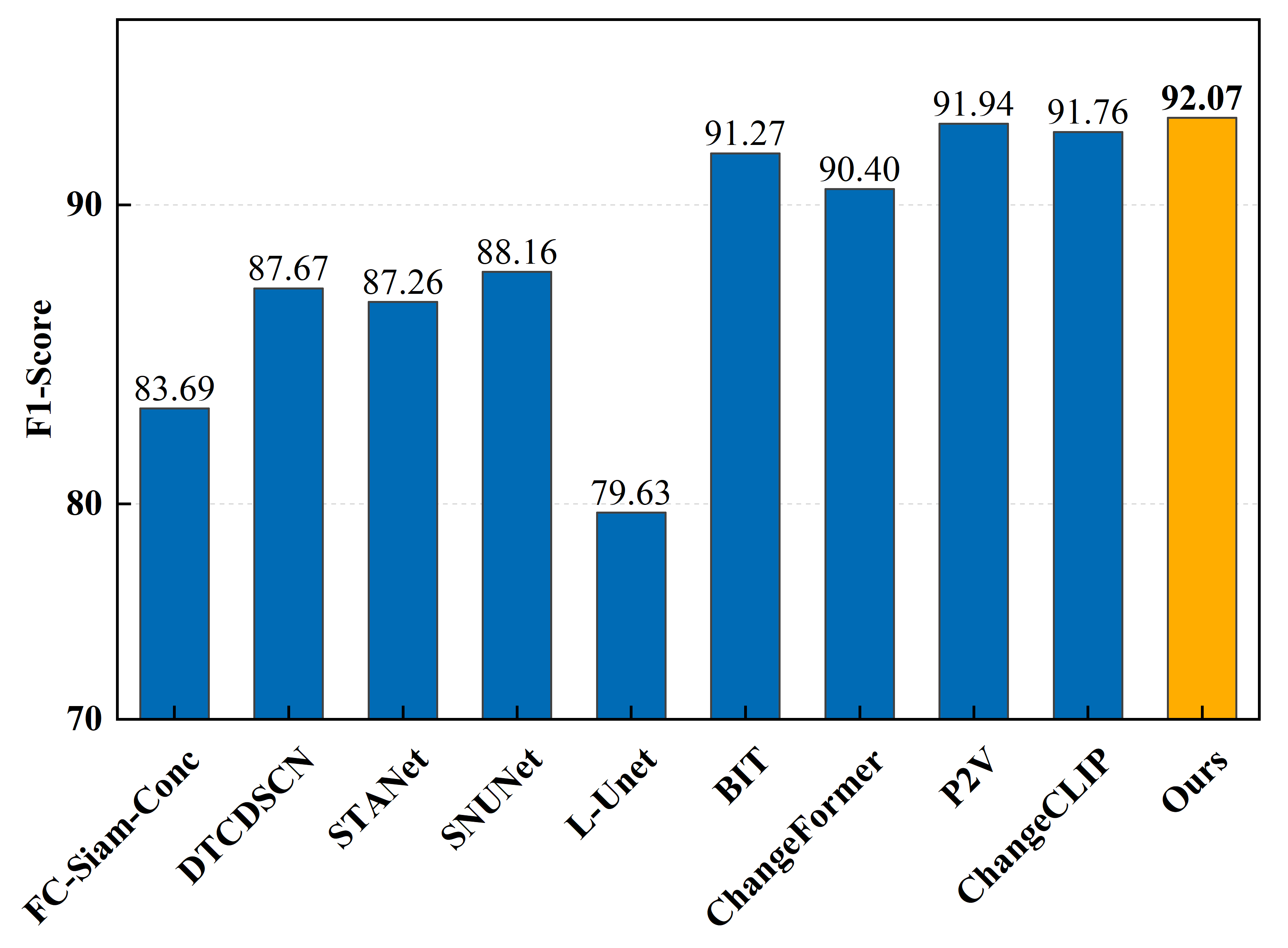}
	\vspace{-1em}
	\caption{Comparison experiment results on LEVIR-CD.}
	\label{LEVIR}
\end{figure}

\begin{figure*}[!t]
	\centering
	\includegraphics[width=7in]{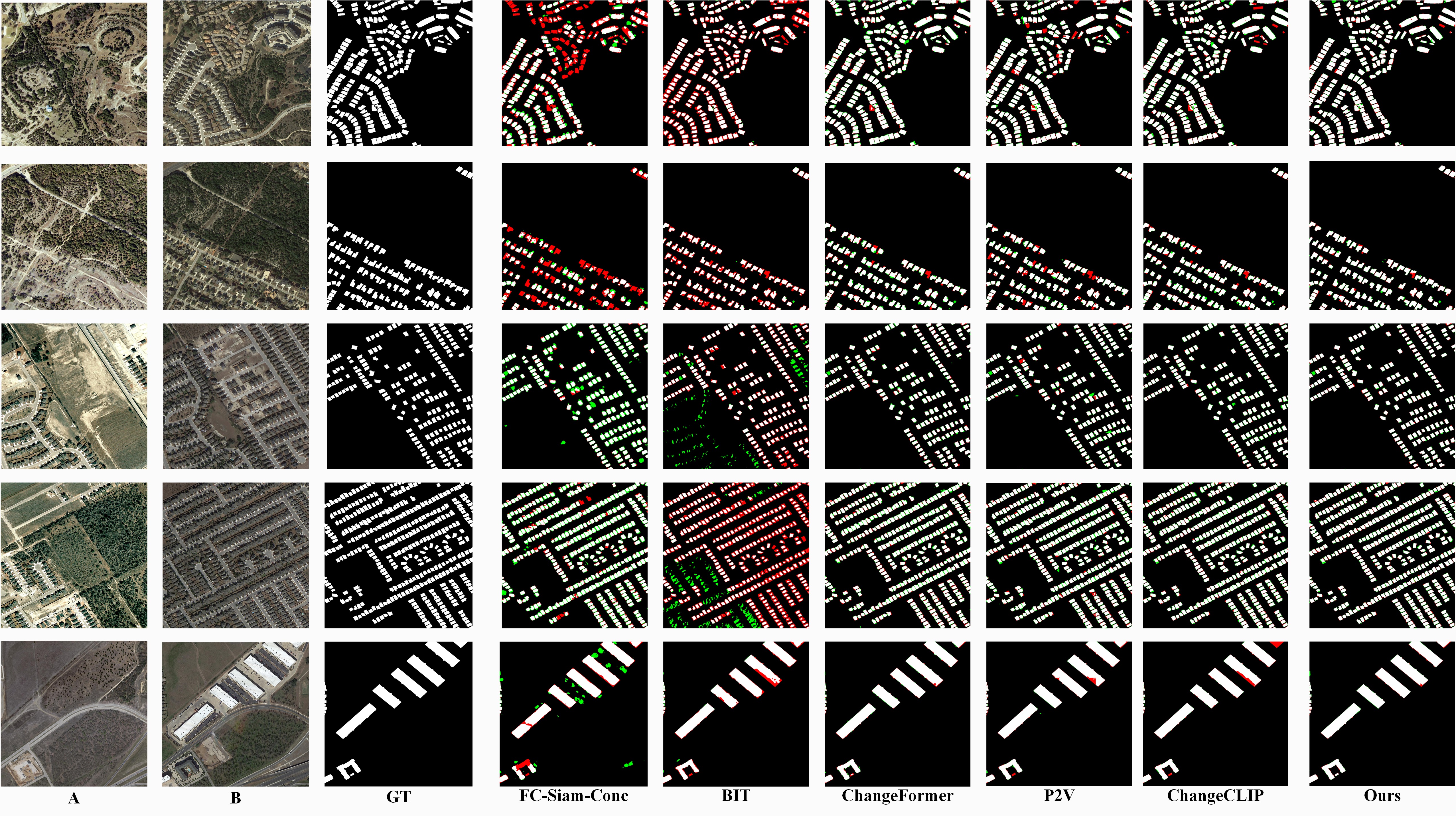}
	\caption{Local visualization of different methods on the LEVIR-CD. True positives (TP) are represented by white pixels, while true negatives (TN) are black. False positives (FP) are green and false negatives (FN) are red.}
	\label{levir_result}
\end{figure*}

The experimental results on the LEVIR-CD dataset are shown in Table \ref{tab3}, with a histgram as shown in Fig. \ref{LEVIR}. Our BGFD achieved 99.20\% Overall Accuracy, 85.30\% IoU, 92.07\% F1-Score, 90.77\% Recall, and 93.41\% Precision. Comparatively, our baseline method, ChangeCLIP, obtained 99.17\% Overall Accuracy, 84.77\% IoU, 91.76\% F1-Score, 90.28\% Recall, and 93.28\% Precision. We observed improvements of +0.03\% in Overall Accuracy, +0.53\% in IoU, +0.31\% in F1-Score, +0.49\% in Recall and +0.13\% in Precision, respectively. Fig. \ref{levir_result} presents the detection results on the LEVIR-CD dataset.

\subsubsection{S2Looking}
\begin{figure}[!t]
	\centering
	\includegraphics[width=3.5in]{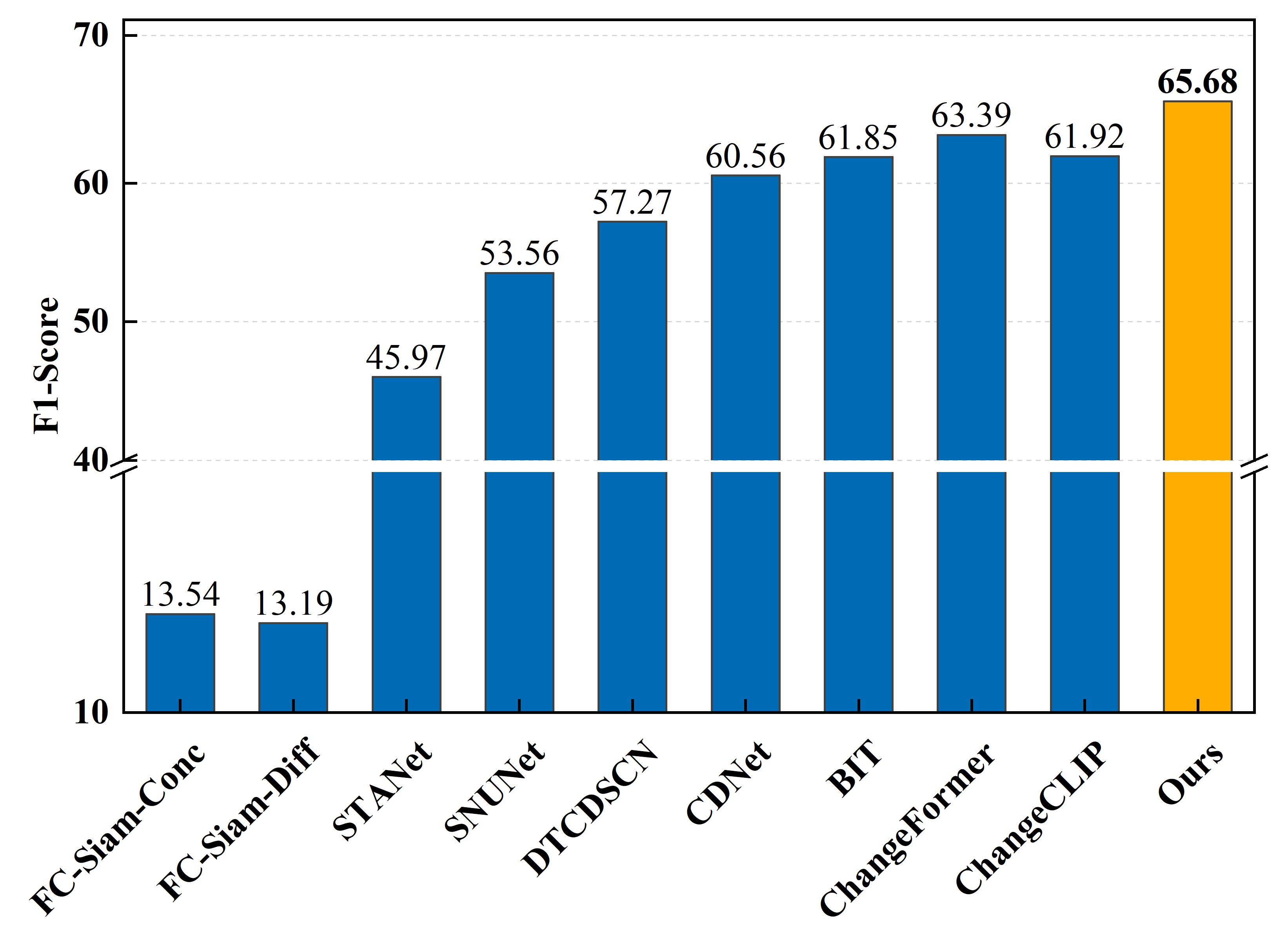}
	\vspace{-1em}
	\caption{Comparison experiment results on S2Looking.}
	\label{S2Looking}
\end{figure}

\begin{figure*}[!t]
	\centering
	\includegraphics[width=7in]{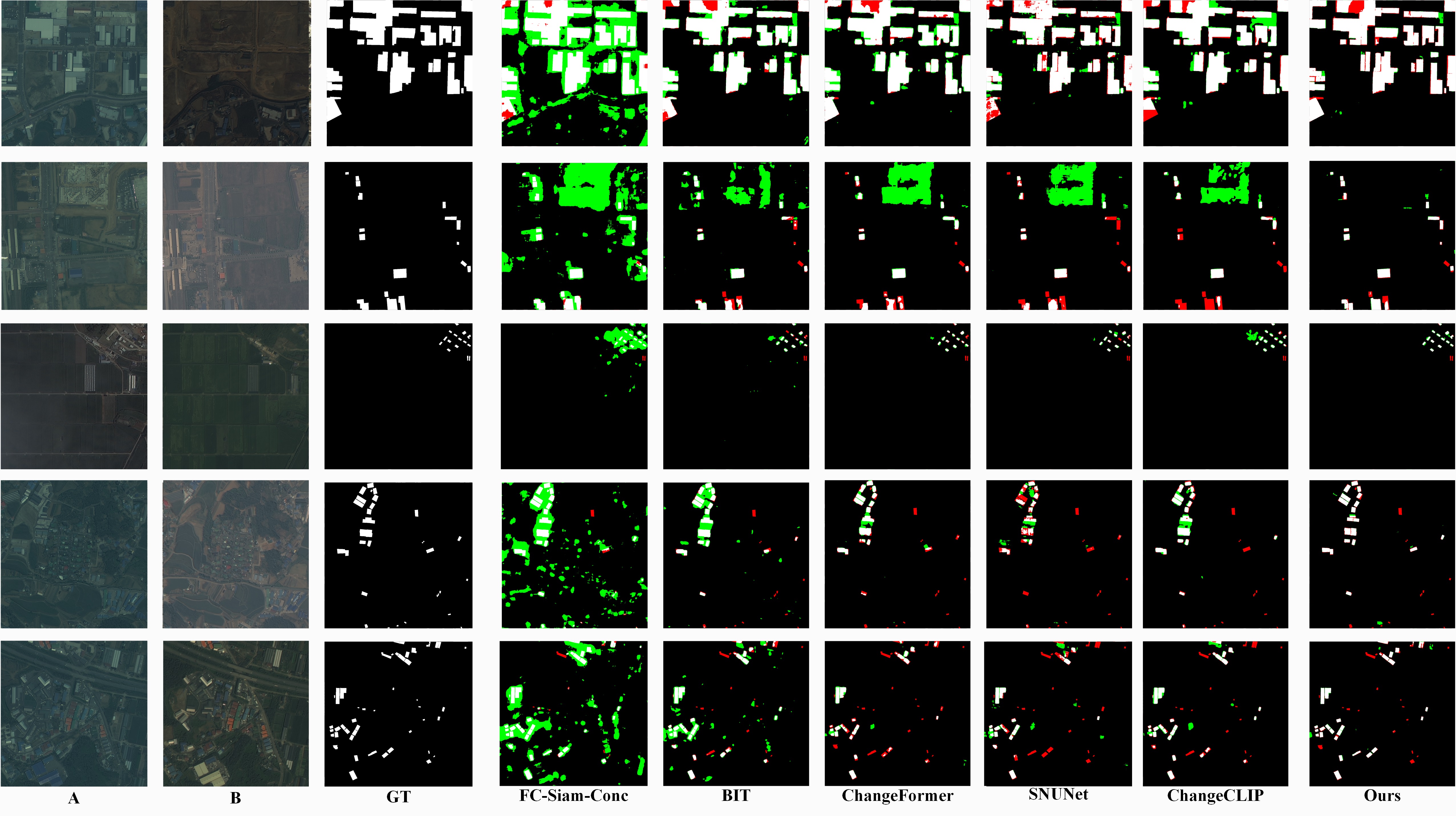}
	\vspace{-1em}
	\caption{Local visualization of different methods on the S2Looking. True positives (TP) are represented by white pixels, while true negatives (TN) are black. False positives (FP) are green and false negatives (FN) are red.}
	\label{s2looking_result}
\end{figure*}

In order to further verify the effectiveness of our network, we selected an additional dataset for experimentation. As most studies using the S2Looking dataset for experiments only compare based on the F1-Score, Recall, and Precision metrics, we conduct comparisons using only these three metrics as well. Table \ref{tab4} and Fig. \ref{S2Looking} presents our experimental results on the S2Looking dataset, where we achieved 65.68\% F1-Score, 58.76\% Recall, and 74.45\% Precision. In comparison, ChangeCLIP scored 61.92\% F1-Score, 53.17\% Recall, and 74.12\% Precision. We observed improvements of +3.76\% in F1-Score, and +5.59\% in Recall and +0.33\% in Precision. Fig. \ref{s2looking_result} depicts the detection result images on the S2Looking dataset.

\begin{table}
	\centering
	\caption{Quantitative Results On The \textbf{S2Looking} Dataset. The Best And Second Results Are Marked In \textbf{Bold} And \uline{Underlined}, Respectively. All These Scores Are Written In Percentage (\%)}
	\begin{tabular}{ccccccc}
		\specialrule{1.2pt}{1pt}{1.5pt}
		\multirow{2}{*}{\textbf{Methods}} & & \multicolumn{5}{c}{\textbf{S2Looking}} \\
		\cmidrule(r){3-7}
		\textbf{} &  & F1 &  & Rec. &  & Prec. \\
		\specialrule{1.2pt}{1pt}{1pt}
		FC-Siam-Conc & & 13.54 & & 18.52 & & 68.27  \\
		FC-Siam-Diff & & 13.19 & & 15.76 & & \textbf{83.29}  \\
		STANet  &	   & 45.97 & & \uline{56.49} & & 38.75  \\
		DTCDSCN & 	   & 57.27 & & 49.16 & & 68.58  \\
		SNUNet  &     & 53.56 & & 49.86 & & 57.84  \\
		CDNet   &     & 60.56 & & 54.93 & & 67.48  \\
		BIT  	 &	   & 61.85 & & 53.85 & & 72.64  \\
		ChangeFormer &  & \uline{63.39} & & 56.13 & & 72.82  \\
		ChangeCLIP & & 61.92 & & 53.17 & & 74.12  \\
		\specialrule{1.2pt}{1pt}{1.5pt}
		\textbf{BGFD} & & \textbf{65.68} & & \textbf{58.76} & & \uline{74.45}   \\
		\Xhline{1.2pt}   
	\end{tabular}
	\label{tab4}
	\vspace{-1em}
\end{table}

\subsection{Ablation Study}
In this section, to validate the effectiveness of the proposed modules, we conducted ablation experiments on each module across four datasets. Subsequently, we further conducted extensive comparative ablation experiments on the DSIFN-CD dataset to verify the rationale and effectiveness of our approach of incorporating distribution into the change detection network. 

\subsubsection{Different Components Analysis}
Initially, the DFC, GNDD, and FDF modules were individually removed from the BGFD, and the performance changes were observed. As FDF was introduced as a module to assist GNDD, we did not use FDF separately but compared the impact of GNDD on network performance with and without FDF. Ablation experiments were conducted on four datasets – LEVIR-CD, DSIFN-CD, S2Looking, and SYSU-CD to validate the effectiveness of the proposed modules. The experimental results are presented in Table \ref{tab5} and Table \ref{tab6}, with F1-Score chosen as the comparative metric due to its comprehensiveness. The performance across the datasets indicates that the addition of each module has a positive impact on the network. Additionally, combining FDF with GNDD yields performance gains, and the optimal performance is achieved when all modules are used in conjunction. 

\begin{figure}[!t]
	\centering
	\includegraphics[width=3.5in]{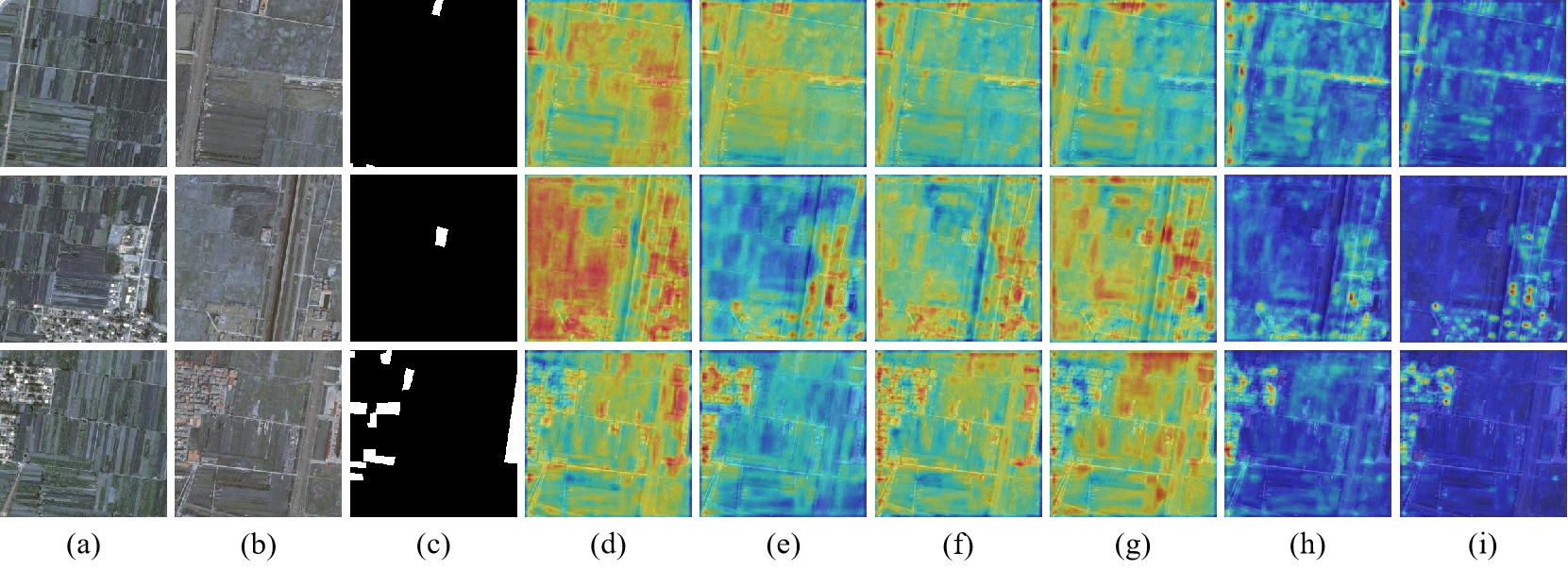}
	\vspace{-1em}
	\caption{Heatmap of visualization results of the ablation study on the DSIFN-CD. (a) ImgA, (b) ImgB, (c) Ground Truth, (d) baseline, (e) GNDD, (f) DFC, (g) DFC + GNDD, (h) GNDD + FDF, (i) BGFD.}
	\label{dsifn_feature_ablation}
\end{figure}

To better demonstrate the effectiveness of the modules, we also compare the visual heatmaps of the network when incorporating different modules, as shown in Fig. \ref{dsifn_feature_ablation}. It can be observed that after integrating the GNDD module, the focus of the model on pseudo changes significantly diminish. With the sole addition of the DFC module, the details of the heatmap become clearer and more comprehensive. When both the GNDD and DFC modules are applied, the heatmap comprehensively exhibits the aforementioned advantages. However, due to insufficient network training stability and the fact that DFC increases the attention of the network on details, the ability to eliminate pseudo changes weakens compared to when only the GNDD module is added. Additionally, the application of both GNDD and FDF demonstrates a noticeable enhancement in the ability to exclude pseudo changes, validating the role of FDF in promoting the effectiveness of GNDD. Lastly, incorporating all modules, the features generated by BGFD not only excel in eliminating pseudo changes but also demonstrate outstanding detail capturing capabilities.

\begin{table}
	\centering
		\caption{F1-Scores Of Ablation Study Of Core Modules Of Our Proposed BGFD On The \textbf{Dsifn-cd} and \textbf{Sysu-cd}.}
		\begin{tabular}{cccc|cc}
			\specialrule{1.2pt}{1pt}{1pt}
			\textbf{Index} & \textbf{DFC} & \textbf{GNDD} & \textbf{FDF} & \textbf{DSIFN} & \textbf{SYSU} \\
			\hline 
			1     & \ding{55}             & \ding{55}              & {\ding{55}}            & 86.10           & 82.98            \\
			2     & \ding{51}            & \ding{55}              & \ding{55}             & 87.30            & 83.16            \\
			3     & \ding{55}             & \ding{51}              & \ding{55}             & 88.05          & 83.31            \\
			4     & \ding{55}             & \ding{51}              & \ding{51}            & 92.03         & 83.38                \\
			5     & \ding{51}             & \ding{51}              & \ding{55}            & 90.76           & 83.59            \\
			6     & \ding{51}             & \ding{51}              & \ding{51}             & \textbf{94.68}        & \textbf{84.26}			  \\
			\Xhline{1.2pt}            
		\end{tabular}
		\vspace{-1em}
		\label{tab5}
\end{table}

\begin{table}
	\centering
		\caption{F1-Scores Of Ablation Study Of Core Modules Of Our Proposed BGFD On The \textbf{Levir-cd} and \textbf{S2Looking}.}
		\begin{tabular}{cccc|cc}
			\specialrule{1.2pt}{1pt}{1pt}
			\textbf{Index} & \textbf{DFC} & \textbf{GNDD} & \textbf{FDF} & \textbf{LEVIR} & \textbf{S2Looking} \\
			\hline 
			1     & \ding{55}             & \ding{55}              & \ding{55}             & 91.76             & 61.92      \\
			2     & \ding{51}             & \ding{55}              & \ding{55}             & 91.87             & 63.05               \\
			3     & \ding{55}             & \ding{51}              & \ding{55}             & 91.87             & 63.17              \\
			4     & \ding{55}             & \ding{51}              & \ding{51}           & 91.94             & 65.03              \\
			5     & \ding{51}             & \ding{51}              & \ding{55}             & 92.01             & 65.13              \\
			6     & \ding{51}             & \ding{51}              & \ding{51}             & \textbf{92.07}             & \textbf{65.68}              \\
			\specialrule{1.2pt}{1pt}{1pt}            
		\end{tabular}
		\vspace{-1em}
		\label{tab6}
\end{table}

\begin{table}
	\renewcommand{\arraystretch}{1.25}
	\centering
		\caption{Ablation Study Of Different Methods To Introduce Distribution On The \textbf{Dsifn-cd}. All These Scores Are Written In Percentage (\%)}
		\begin{tabular}{c|ccccc}
			\specialrule{1.2pt}{1pt}{1pt}
			\textbf{Methods}  & \textbf{OA} & \textbf{IoU} & \textbf{F1} & \textbf{Rec.} & \textbf{Prec.} \\
			\hline
			Baseline & 95.13       & 75.59        & 86.10       & 88.68        & 83.66         \\
			As Loss        & 93.95       & 71.30        & 83.24       & 88.44        & 78.62         \\
			As DI        & 94.42       & 72.91        & 84.33       & 88.32        & 80.69         \\
			Noise Swap        & 95.17       & 75.40        & 85.98       & 87.09        & 84.89         \\
			GNDD        & \textbf{99.19}       & \textbf{85.02}        & \textbf{91.87}       & \textbf{90.49}        & \textbf{93.30}    		\\
			\specialrule{1.2pt}{1pt}{1pt}    
		\end{tabular}
		\vspace{-1em}
	\label{tab7}
\end{table}

\subsubsection{Different methods to introduce distribution}
In our experiments, we explored various methods to integrate the concept of distribution into the change detection network. We conducted the following comparative experiments:
\begin{itemize}
	\item {\textbf{As Loss}: Approximating the Gaussian distribution of bi-temporal images separately, calculating the Wasserstein distance between the two distributions, and using this distance as a loss to constrain network training.}
	\item{\textbf{As DI}: Approximating the Gaussian distribution of bi-temporal images separately, calculating the Wasserstein distance between the two distributions, and using the distance images of each channel as a form of attention, weighted and multiplied to the original image stack, considered as a new difference image.}
	\item{\textbf{Noise swap}: Approximating the Gaussian distribution of bi-temporal images separately and generating corresponding noise. However, distinct from our employed method, we applied the noise to the feature map of the other temporal image. In other words, noise generated based on the distribution from time A was added to the feature map of time B.}
	\item{\textbf{GNDD}: The approach which we applied in this paper.}
\end{itemize}

It is worth noting that this ablation study was conducted on the baseline model without incorporating DFC and FDF. The results of the comparative experiments are presented in Table \ref{tab7}. From the results, it can be seen that the method we ultimately adopted is the most effective among the various approaches mentioned above. At the same time, it is also evident that inappropriate ways of introducing distribution may even lead to a decrease in network performance.

\section{CONCLUSION}
The BGFD proposed in this paper fundamentally addresses the issue of pseudo changes caused by differences in domain information from a distribution perspective. It also compensates for the loss and contamination of detail features during the upsampling process, overall enhancing the detection capability and robustness of our network. The Gaussian noise domain disturbance module leverages the statistical features of images to characterize domain information, perturbing the network to learn redundant domain information, thereby enhancing the domain robustness of the network. The feature dependency facilitation module enhances the adaptability of the network and constrains the training process to ensure the acquisition of essential domain information. The detail feature compensation module enhances the detail information and refines global information of small-scale feature maps respectively, improving the model ability to detect details. Our model has achieved state-of-the-art results on the DSIFN-CD, SYSU-CD, LEVIR-CD and S2Looking publicly available datasets. 

However, in cases where our model encounters image pairs with significant differences in perspective or severe pixel misalignment, misaligned regions may be interpreted by the network as representing a different domain of information. Consequently, the network may overlook these areas of change, leading to suboptimal detection results. We will further investigate these limitations in our future research. 

\normalem
\bibliographystyle{IEEEtran}
\bibliography{mybibfile}

\newpage
\newpage

\end{document}